\documentclass[Afour,sageh,times]{sagej}

\usepackage{moreverb,url}

\usepackage{bbding}
\usepackage{moreverb,url}
\usepackage{url}
\usepackage{tablefootnote}
\usepackage[linktoc=all,colorlinks,bookmarksopen,bookmarksnumbered,citecolor=blue,urlcolor=blue]{hyperref}
\usepackage{balance}
\usepackage[table]{xcolor}
\usepackage{graphicx}
\usepackage{amsfonts}
\usepackage{fancyhdr}
\usepackage{comment}
\usepackage{times}
\usepackage{amsmath}
\usepackage{changepage}
\usepackage{amssymb}
\usepackage{enumerate}
\usepackage{bm}
\usepackage{calligra}
\usepackage{multirow}
\usepackage{tabularx}
\usepackage{booktabs}
\usepackage{mathtools}
\usepackage{arydshln}
\usepackage{latexsym} 
\usepackage{amssymb}
\usepackage{color}
\usepackage[linesnumbered,ruled,vlined]{algorithm2e}
\usepackage{ulem}
\usepackage{float}
\usepackage{tabstackengine}
\usepackage{siunitx}
\usepackage{colortbl}
\usepackage{hhline}
\usepackage{bookmark}
\usepackage{dirtree}
\usepackage{subfigure}
\usepackage{wasysym}
\usepackage{enumitem}
\usepackage{overpic}
\usepackage{pdflscape}
\usepackage{afterpage}

\usepackage{makecell}
\usepackage{gensymb}

\newcommand\BibTeX{{\rmfamily B\kern-.05em \textsc{i\kern-.025em b}\kern-.08em
T\kern-.1667em\lower.7ex\hbox{E}\kern-.125emX}}

\def\volumeyear{2025}

\begin{document}

\runninghead{Ye et al.}

\title{TPT-Bench: A Large-Scale, Long-Term and Robot-Egocentric Dataset for Benchmarking Target Person Tracking}

\author{Hanjing Ye\affilnum{1}, Yu Zhan\affilnum{1}, Weixi Situ\affilnum{1}, Guangcheng Chen\affilnum{1}, Jingwen Yu\affilnum{1,4}, Ziqi Zhao\affilnum{3}, Kuanqi Cai\affilnum{2}, Arash Ajoudani\affilnum{2} and Hong Zhang$^{1*}$}

\affiliation{\affilnum{1}Robotics and Computer Vision Laboratory, Southern University of Science and Technology, Shenzhen, China\\
\affilnum{2}Human-Robot Interfaces and Interaction Laboratory, Italian Institute of Technology, Geona, Italy\\
\affilnum{3}Robotics Perception and Intelligence Laboratory, Southern University of Science and Technology, Shenzhen, China\\
\affilnum{4}Chen Kar-Shun Robotics Institute, Hong Kong University of Science and Technology, Hong Kong SAR, China}

\corrauth{Hong Zhang, hzhang@sustech.edu.cn}

\begin{abstract}
Tracking a target person from robot-egocentric views is crucial for developing autonomous robots that provide continuous personalized assistance or collaboration in Human-Robot Interaction (HRI) and Embodied AI. However, most existing target person tracking (TPT) benchmarks are limited to controlled laboratory environments with few distractions, clean backgrounds, and short-term occlusions. In this paper, we introduce a large-scale dataset designed for TPT in crowded and unstructured environments, demonstrated through a robot-person following task. The dataset is collected by a human pushing a sensor-equipped cart while following a target person, capturing human-like following behavior and emphasizing long-term tracking challenges, including frequent occlusions and the need for re-identification from numerous pedestrians. It includes multi-modal data streams, including odometry, 3D LiDAR, IMU, panoramic images, and RGB-D images, along with exhaustively annotated 2D bounding boxes of the target person across 48 sequences, both indoors and outdoors. Using this dataset and visual annotations, we perform extensive experiments with existing SOTA TPT methods, offering a thorough analysis of their limitations and suggesting future research directions. Our dataset, code and video are available at \href{https://medlartea.github.io/tpt-bench/}{https://medlartea.github.io/tpt-bench/.}
\end{abstract}

\maketitle
\renewcommand{\thefootnote}{\arabic{footnote}}

\begin{figure*}
    \centering
    \includegraphics[width=\textwidth, trim=0 2 5 2, clip]{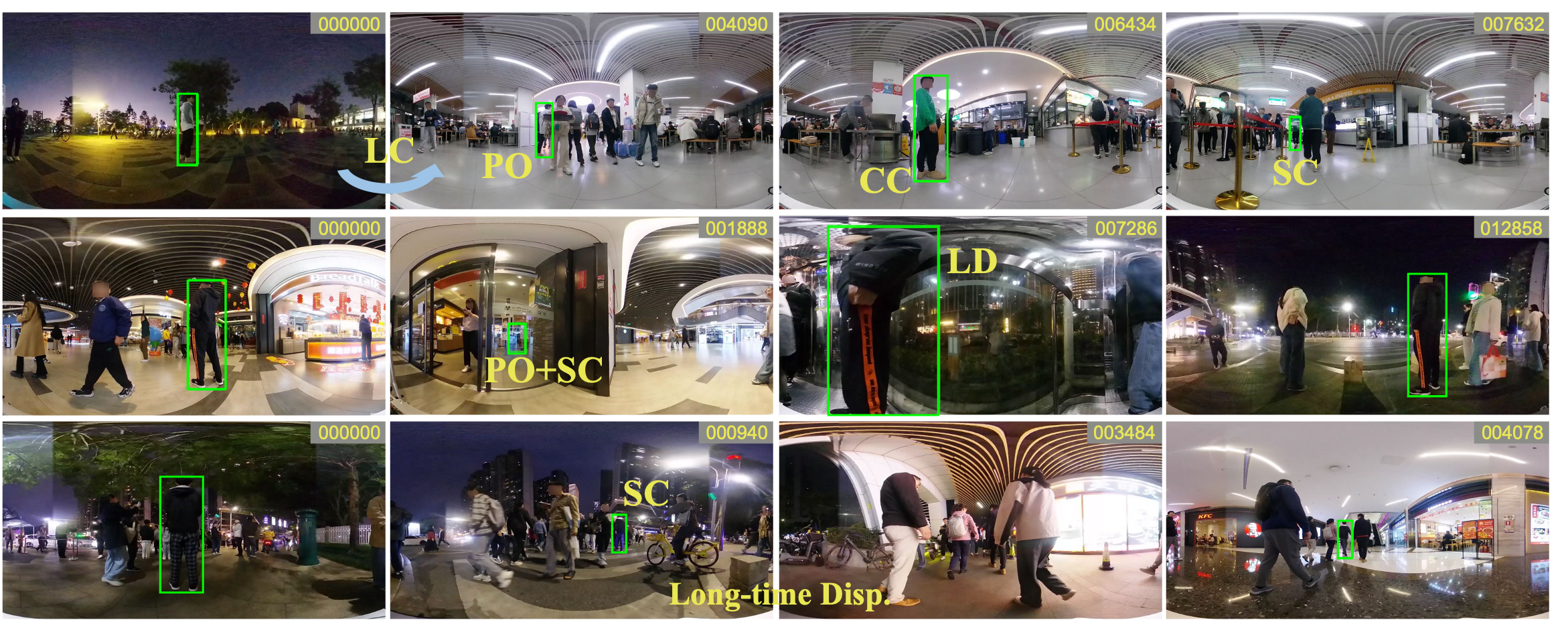}
    \caption{\textbf{Example robot-egocentric visual sequences and annotations of the TPT-Bench dataset.} This dataset is collected by a human pushing a sensor-equipped cart following a target person in diverse crowded environments, e.g., canteens, crossroads, classrooms, markets, etc. In addition to challenges involved in general visual tracking benchmarks, e.g., partial occlusion (PO), large distortion (LD), clothes change (CC), scale change (SC), lighting change (LC) and motion blur, the TPT-Bench emphasizes long-term tracking challenges, including frequent target disappearance (Disp.) and the need for target re-identification among numerous pedestrians. The upper-right number denotes the frame number.}
    \label{fig:teaserImage}
\end{figure*}

\section{Introduction}
To effectively perform Human-Robot Interaction (HRI) and Embodied AI tasks in environments with people, robots must recognize their human counterparts and adapt their actions to provide personalized assistance. The ability to continuously recognize and track a target person, known as \textbf{target person tracking (TPT)}, is an egocentric perceptual task related to navigation that has recently gained significant attention in the robotics and HRI communities~\citep{islam2019person,singamaneni2024survey,zhanghao2020ijrr,ye2023robot,ye2024person,rollo2024icra,eirale2025human,koide2020monocular,zhao2024human}. While these frameworks demonstrate efficient identification, they are often evaluated in controlled laboratory settings with minimal distractions, clean backgrounds, and short tracking durations.

Datasets for evaluating TPT are limited. \cite{chen2017crv} introduced a public TPT dataset featuring challenging scenarios such as short-term occlusions by rapid multi-person crossings, illumination changes, and appearance variations. However, it lacks more challenging scenarios, such as occlusions and distractions caused by people with similar appearances. To address this gap, some methods have proposed custom-built datasets~\citep{ye2024person,rollo2024icra,koide2020monocular,gross2017roreas,eisenbach2023little}, but their experimental settings are constrained, typically involving only a few distractors and being collected in controlled laboratory environments with clean backgrounds. Such small-scale and structured environments limit the development of TPT. 

TPT focuses on continuous tracking of a target person, with related datasets including multiple-object tracking (MOT)~\citep{milan2016mot16, manen2017pathtrack, dendorfer2020mot20} and single-object tracking (SOT)~\citep{WuLimYang13, muller2018trackingnet, fan2021lasot}. While SOT datasets address challenges like occlusion and pose changes, they are designed for generic object tracking and lack a moving egocentric viewpoint for performing robot companionship in crowded and unstructured environments. Such a scenario involves uncontrolled natural conditions, especially frequent occlusion and obstruction. In addition, SOT datasets offer limited trajectories for people and short-duration recordings. MOT datasets focus on human trajectories in crowded scenes, but the recoding time for individual subjects is still limited.

To address these gaps, we propose a TPT dataset that captures diverse scenarios from a moving egocentric perspective in crowded environments, highlighting the challenges of frequent occlusion and obstruction from numerous pedestrians. The dataset focuses on a realistic HRI task: robot person following~\citep{islam2019person} or assistive socially-aware navigation~\citep{singamaneni2024survey}. It is collected by a human pushing a sensor-equipped cart while following a target person, as shown in Figure~\ref{fig:sensorSuit}. This human-human interaction behavior offers valuable insights for developing human-like person-following methods~\citep{karunarathne2018model}. The dataset comprises 5.3 hours of multi-modal recordings, including robot odometry, 3D LiDAR, panoptic and RGB-D images, as well as IMU streams. In addition to the raw data, we provide frame-level 2D bounding box annotations of the target person, totaling 571,982 frames. Collected across 48 sequences in densely populated environments---such as schools, food stores, markets, plazas, and metro stations---the dataset enables long-term tracking episodes, with each sequence averaging 397.2 seconds in duration.

Building on this well-annotated dataset, we introduce \textbf{TPT-Bench}, a unified benchmark for evaluating various TPT algorithms. In summary, the main contributions of this paper are: 
\begin{itemize}
    \item A large-scale, long-term, multi-modal, and robot-egocentric TPT dataset highlighting long-term tracking challenges of frequent target disappearance in diverse, crowded environments.
    \item An evaluation framework that measures a TPT algorithm's ability to identify and track the true target person with accuracy and confidence.
    \item A systematic analysis and comparison of existing TPT algorithms on the TPT-Bench, LaSOT~\citep{fan2021lasot}, and Chen's dataset~\citep{chen2017crv}, demonstrating the TPT-Bench is more challenging with the objective of long-term tracking.
\end{itemize}

\section{Related Work}
In this section, we review related literature with a focus on two key areas: 1) \textbf{Target Person Tracking}, illustrating existing works on target person tracking; and 2) \textbf{Datasets for TPT, SOT and MOT}, explaining why existing datasets are inadequate for developing TPT algorithms.

\begin{table*}[t]
    \centering
    \caption{Public TPT datasets and related person-tracking datasets, including MOT and SOT datasets. For SOT datasets, we focus solely on video sequences involving the person category. In the case of MOT datasets, which feature trajectories of multiple individuals, we report attributes for each person as a sequence across all videos. Therefore, MOT20~\citep{dendorfer2020mot20} and JRDB~\citep{martin2021jrdb} report 2257$^*$ and 1805$^*$ instances, respectively, but actually encompass only 4 and 27 scenarios. Typically, JRDB has only 27 scenarios labels open sourced, and only 16 contain robot-moving behavior. Comparison of long-term properties: the number of sequences (\textbf{\#sequences}), the number of scenarios (\textbf{\#scenarios}), the average length of a person (\textbf{APL}, s), the average disappearance length (\textbf{ADL}, s), the average distractors per second in the disappearance interval (\textbf{ADNDS}), the number of target disappearances (\textbf{DSP}), the average number of disappearances in sequence (\textbf{ADN})$=$\textbf{DSP}$/$\textbf{\#sequences}.}
    \scalebox{0.95}{
    \begin{tabular}{l|cccccccc}
        \toprule
        \bf Benchmark &\bf Task &\bf \makecell{\#sequences} &\bf \makecell{APL (s)} &\bf \makecell{ADL (s)} & \bf ADNDS &\bf \makecell{DSP} &\bf \makecell{ADN} &\bf \makecell{Robot \\ Egocentric}\\
        \midrule
        \midrule
        \bf LaSOT~\citep{fan2021lasot} &SOT &20 &106.8 &0.3 &7.8 &31 &1.6 &\XSolidBrush \\
        \bf LTB50~\citep{lukezivc2020performance} &SOT &20 &106.3 &9.1 &0.8 &97 &4.9 &\XSolidBrush\\
        \midrule
        \bf MOT20~\citep{dendorfer2020mot20} &MOT &2257$^*$ &20.5 &1.5 &126 &1617 &0.7 &\XSolidBrush\\
        \bf JRDB~\citep{martin2021jrdb} &MOT &1805$^*$ &65.7 &11.5 &47 &7265 &4.0 &\Checkmark\\
        \midrule
        \bf Chen's dataset~\citep{chen2017crv} &TPT &11 &340.5 &1.8 &3 &4 &0.4 &\Checkmark\\
        \rowcolor{gray!25}
        \bf TPT-Bench (ours) &TPT &{48} &{397.2} &{65.3} &{9.6} &{2177} &{45.4} &\Checkmark\\
        \bottomrule
    \end{tabular}}
    \label{tab:datasets}
\end{table*}

\subsection{Target Person Tracking}
Target person tracking aims to identify and track a specific person, which is crucial for personalized HRI scenarios where the robot needs to engage with its user over extended periods~\citep{islam2019person,singamaneni2024survey}. Earlier TPT approaches often relied on visual tracking techniques, such as MOT or SOT, to track the target person.
SOT-assisted methods initially identify the target person by selecting their bounding box in the first frame, then use an appearance model to search for the target within a nearby region~\citep{chen2017crv, zhang2019vision}. SOT techniques~\citep{javed2022visual} aim to track an arbitrary object by simultaneously classifying and localizing it within a search region. However, due to the lack of prior knowledge about human appearance (e.g., from pre-trained models of people detection and person re-identification), these methods can easily track a wrong person in crowded environments, particularly after partial or total occlusions.

Conversely, MOT techniques~\citep{luo2021multiple} assume prior knowledge of the tracked object, typically people. MOT techniques usually detect individuals using well-trained object detectors and track multiple people within the sensor's field of view (FoV). Based on these well-tracked people, MOT-assisted TPT methods then identify the target person by selecting their identity ~\citep{zhang2015icra, repiso2020adaptive}. Such a solution is robust to short-term occlusions but still struggles with long-term occlusions since they often lack a target re-identification (ReID) capability after occlusions.

To address these gaps, recent TPT methods typically combine MOT with a target ReID module. The MOT front-end serves two main purposes: 1) tracking people’s trajectories is essential for socially-aware path planning~\citep{repiso2020adaptive, morales2017social, triebel2016spencer}, which helps the robot assist the target person while respecting proxemics with others, and 2) by re-identifying the target from tracked individuals, established person ReID techniques~\citep{ye2021deep} can be utilized in the back-end module to recognize the target person after partial or total occlusions. With the advancement of deep learning, deep-learning-based (DL-based) ReID features have become dominant due to their robustness to viewpoint and illumination changes~\citep{ye2021deep}.

To leverage the power of DL-based features for target ReID, several approaches have been proposed~\citep{rollo2024icra, koide2020monocular, ye2024person}. For example, \cite{koide2020monocular} introduced an online boosting classifier using low-level CNN features, while \cite{rollo2024icra} represented the target feature as a distribution and updated it continuously with newly archived features using a damped exponential moving average. Moreover, \cite{ye2024person} observed a domain shift in pre-trained ReID features—those from video-surveillance data often fail to generalize to practical person-following environments. To address this, they proposed using online continual learning techniques to optimize ReID features by collecting valuable observations.

This paper evaluates the most recent target-ReID methods~\citep{rollo2024icra,ye2024person} and popular SOT trackers on TPT-Bench, as well as other related datasets like LaSOT~\citep{fan2021lasot} and Chen's dataset~\citep{chen2017crv}, which track a target person over extended periods. It highlights the more challenging long-term tracking scenarios presented by TPT-Bench.

\subsection{Datasets related to Target Person Tracking}
\subsubsection{Datasets for Target Person Tracking}
Datasets for evaluating TPT are limited. The only public dataset~\citep{chen2017icvs} is collected by teleoperating the robot following the target person, involving challenging scenarios such as quick multi-people crossings, illumination changes, and appearance variations. However, this dataset lacks challenging scenarios requiring person ReID, such as occlusion and similar appearances of distracting people. To mitigate this limitation, some methods propose custom-built datasets~\citep{ye2024person,rollo2024icra,koide2020monocular}, but their experimental scenarios are limited, involving only a few distracting people.

Moreover, ROREAS~\citep{gross2017roreas} and ROREAS+~\citep{eisenbach2023little} propose TPT datasets involving scenarios of stroke patient assistance in their daily rehabilitation. Although these include challenges such as frequent occlusions by multiple people and lighting changes, their scenarios are limited to stroke patient assistance and are private due to data protection laws.

\subsubsection{Datasets for SOT and MOT}
Since TPT focuses on continuous online tracking of a target person, related datasets include MOT datasets~\citep{milan2016mot16, manen2017pathtrack, dendorfer2020mot20} and SOT datasets~\citep{WuLimYang13, muller2018trackingnet, fan2021lasot}. SOT datasets, such as LaSOT~\citep{fan2021lasot} and TrackingNet~\citep{muller2018trackingnet}, focus on tracking arbitrary objects, including people, across various scenarios like dance, surveillance, and sports. While these datasets present challenges like occlusion, cluttered backgrounds, and pose variations, they lack robot-egocentric sequences designed for personalized robot assistance. In such scenarios, the tracked person may frequently be occluded or disappear for long periods. Additionally, the robot's perspective differs from that of surveillance or handheld cameras in that people are often partially observed with body parts, making it more challenging to track the person.

MOT datasets, such as MOT20~\citep{dendorfer2020mot20}, are designed for multi-person tracking in crowded environments with frequent occlusions. However, they focus on surveillance footage, where occlusion patterns differ significantly from those seen in robot observations. These datasets also provide limited long-term tracking data for a target person. The JRDB dataset~\citep{martin2021jrdb}, the most relevant to our work, includes egocentric data with 3D LiDAR, IMU and visual data, collected by a teleoperated robot in crowded environments. However, it does not represent a long-term tracking scenario in the context of HRI. Moreover, it assumes people remain within the camera's field of view (FoV) and does not address the need to re-identify individuals when they reappear after being occluded.

A quantitative comparison of person-tracking datasets is presented in Table~\ref{tab:datasets}, with most metrics derived from a long-term tracking benchmark~\citep{lukezivc2020performance}. The TPT-Bench captures long-term robot companionship from an egocentric perspective in crowded, unstructured environments, resulting in more occlusions and longer disappearance periods compared to existing person-tracking datasets. Additionally, when a target reappears, its appearance may differ from its previous state in terms of pose, lighting, and scale, as shown in Figure~\ref{fig:teaserImage}, which presents additional challenges for target ReID. These difficulties are demonstrated in the experiments through a comprehensive comparison of current TPT algorithms on the TPT-Bench.

\section{The TPT-Bench Dataset}
In this section, we first describe the data collection platform (Sec.~\ref{sec:platform}) and introduce the scenarios (Sec.~\ref{sec:scenarios}) collected in this dataset. We then describe the annotation process and standards (Sec.~\ref{sec:label}) and provide dataset statistics (Sec.~\ref{sec:statistics}) of the whole dataset, highlighting unique challenges present by the proposed dataset. In the end, we introduce the dataset organization (Sec.~\ref{sec:dataOrg}) and the development tools (Sec.~\ref{sec:develTools}).

\subsection{Collecting Data}

\subsubsection{Platform} \label{sec:platform}
\begin{figure}[t]
        \centering
        \includegraphics[width=\linewidth]{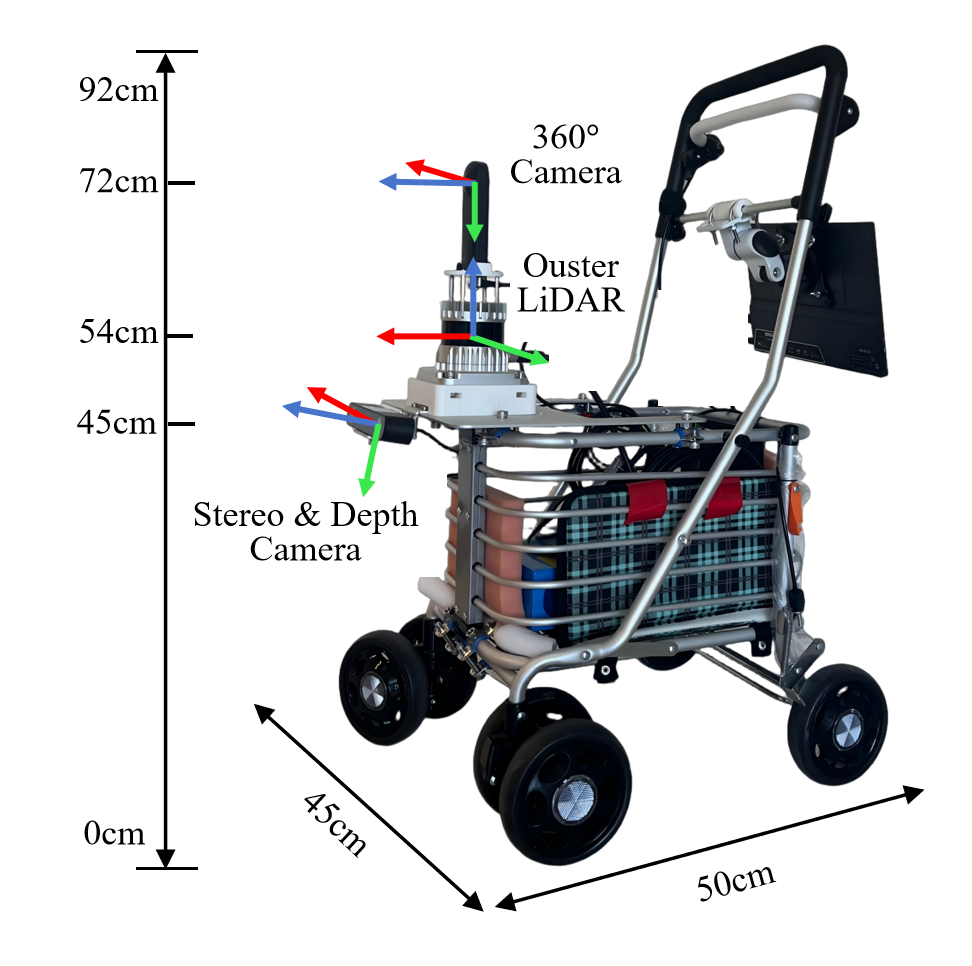}
\caption{\textbf{The platform and sensor suite.} A panoramic camera, depth camera, and 3-D LiDAR are fixed on a push-cart in positions that mirror the typical sensor layout of most mobile robots.}
\label{fig:sensorSuit}
\end{figure}

To facilitate the collection of multi-sensor and egocentric data under robot kinematic constraints for non-robotics experts, we have developed a data collection platform illustrated in Figure~\ref{fig:sensorSuit}. This platform consists of a pushing cart equipped with a panoramic camera, a stereo camera, and a 3D LiDAR, operating under differential wheel constraints. The sensors are mounted at typical heights for mobile robots: the panoramic camera is placed approximately 72 cm above the ground, the 3D LiDAR at about 54 cm, and the stereo camera at approximately 45 cm with an elevation angle of around 30$^{\degree}$. The extrinsic parameters of these sensors are well-calibrated and provided. Based on the designed platform, we collect multi-sensor data. Below, we describe the characteristics of each sensor:
\begin{itemize} 
\item \textbf{Panoramic Camera}: A RICOH Theta Z1 360$^{\degree}$ camera captures images using dual fisheye lenses. It live streams equirectangular projection images after built-in undistortion and stitching, producing RGB images with a resolution of 1920$\times$960 at 30 Hz. 
\item \textbf{Depth Camera}: A ZED2 camera captures RGB-D images at a resolution of 640$\times$480 and operates at 30 Hz. 
\item \textbf{3D LiDAR}: An Ouster OS1-64 LiDAR generates point clouds with a resolution of 64$\times$2048 at 10 Hz. 
\item \textbf{IMU}: Inertial measurement unit data is recorded from the built-in IMU of the Ouster LiDAR. 
\item \textbf{Odometry / Actions}: Similar to SCAND~\citep{karnan2022socially} and MuSoHu~\citep{nguyen2023toward}, we record visual-inertial odometry provided by the ZED 2 camera. This information can be decomposed into linear and angular velocities as the training data for learning-based navigation methods~\citep{nguyen2023toward}.
\end{itemize}

Although our dataset includes multi-sensor data, we have annotated only the images from the panoramic camera to assess TPT performance using vision-based trackers. We will release all data in \textit{ROSBAG} format, enabling future research on multi-modal TPT combining visual and depth modalities, natural following through imitation learning with multiple modalities and odometry, and more. Our data collection approach mirrors a human follower's intention without delay, and all sensors are mounted similarly to the locations of most mobile robots. As such, the collected data can be directly used to train end-to-end networks for performing socially aware companionship, as demonstrated by recent advancements in learning-based social navigation~\citep{karnan2022socially,nguyen2023toward}.

\begin{figure}[t]
        \centering
        \includegraphics[width=\linewidth]{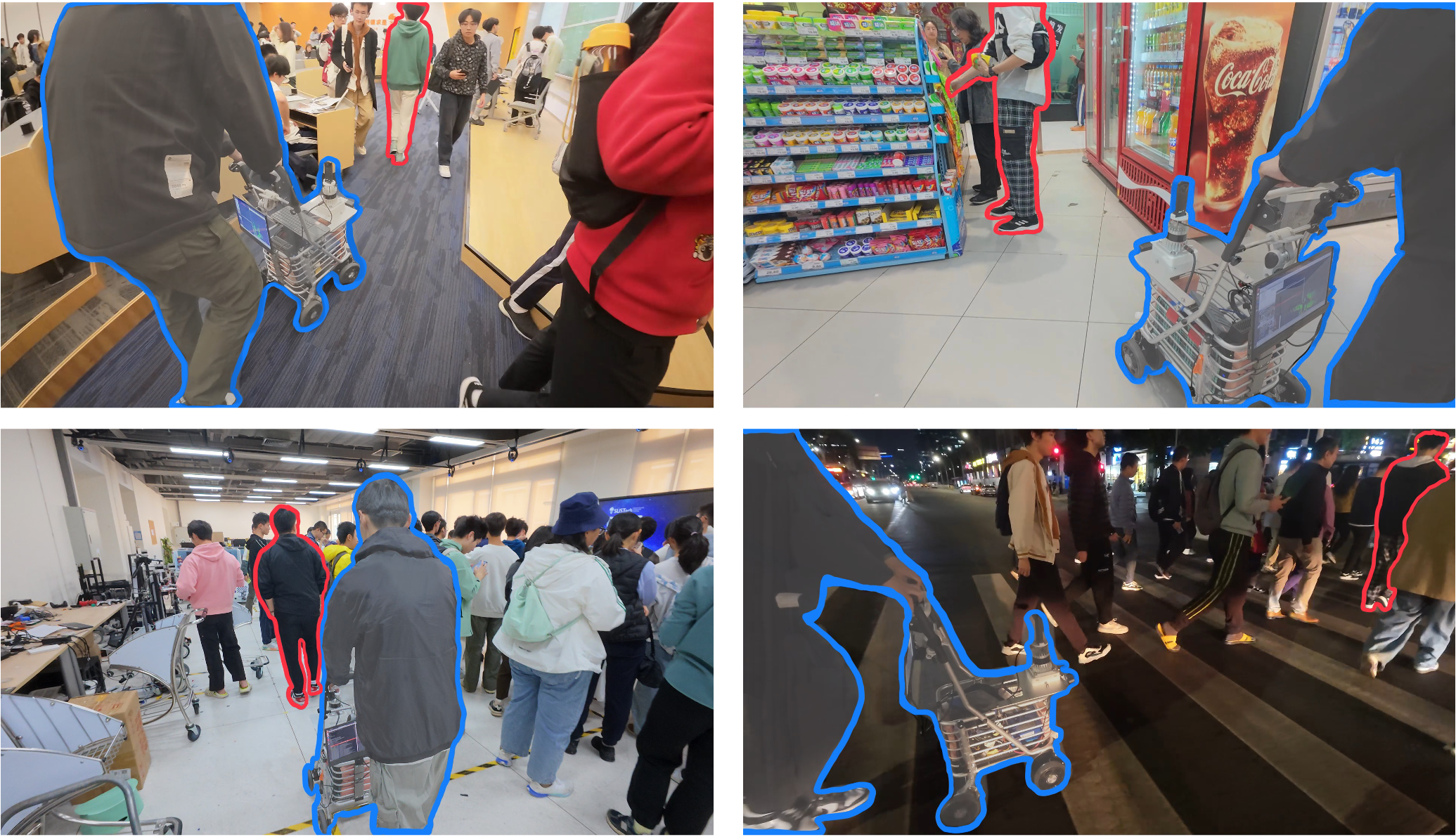}
\caption{\textbf{Third-view examples of dataset collection procedure.} The dataset is demonstrated by a human pushing a sensor-equipped pushcart and following a target person in crowded environments indoors and outdoors. Blue represents the follower, and red represents the target person.}
\label{fig:thirdViews}
\end{figure}

\begin{figure*}[h]
    \centering
    \includegraphics[width=\linewidth]{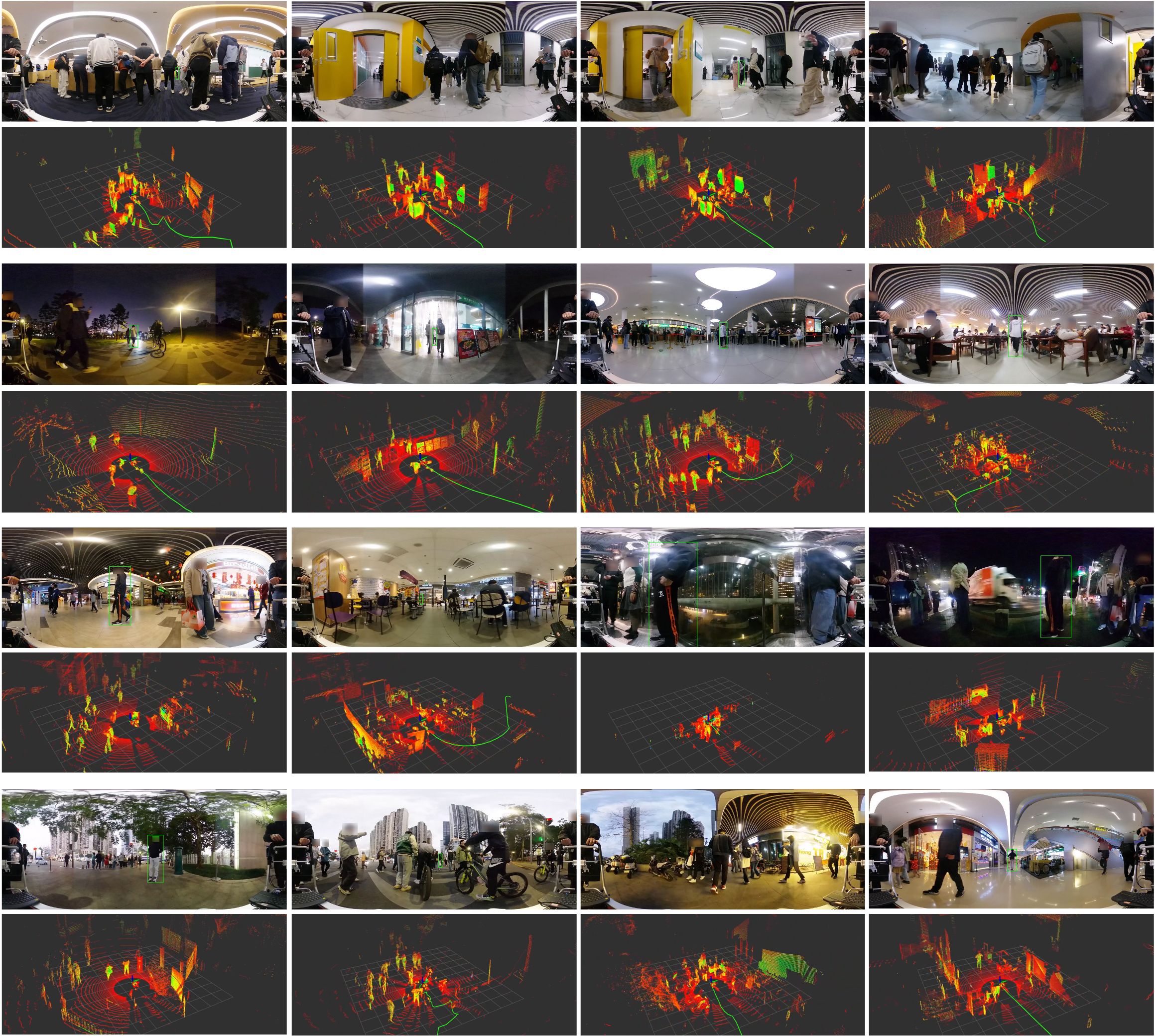}
    \caption{\textbf{Four example sequences from the TPT-Bench dataset.} The top section displays the panoramic images and their corresponding visual annotations, while the bottom section presents the point cloud visualization captured by the 3D LiDAR. The green trajectory in the point cloud represents the robot's path, consisting of the most recent 20 waypoints recorded by the ZED2 VIO. The scenarios, listed from top to bottom, are: ``Teaching Building,'' ``Open Outdoor Area in School and Canteen,'' ``Market, Elevator, and Crossing Roads,'' and ``Crossing Roads, Sidewalk, and Market.''}
    \label{fig:datasetExamples}
\end{figure*}

\subsubsection{Scenarios} \label{sec:scenarios}
The sensor suite described in the previous section is designed to be user-friendly, ensuring accessibility even for those without expertise in robotics. We collected the dataset both on campus and in nearby public areas, focusing on realistic and challenging scenarios to assess the long-term tracking capabilities of TPT algorithms. This dataset highlights challenges such as partial and full occlusions in crowded environments, as well as variations of lighting, pose and scale.

To capture these long-term tracking challenges, we conducted person-following tasks across various indoor and outdoor locations during peak times, such as after classes, during lunch, and at dinner hours. The environments included canteens, laboratories, offices, university building halls, corridors, classrooms, supermarkets, roadside markets, crosswalks, sidewalks, metro stations, and more. Nine students (three females and six males) were recruited as target humans, navigating through these crowded environments and encountering issues like motion blur, varying lighting conditions, changing scales, and alterations in pose and clothing. These students also took on the role of followers, operating the sensor-equipped cart while adhering to social norms, ensuring they did not interrupt the leader or other pedestrians. Additionally, we occasionally carried out person-search tasks~\citep{search2014goldhoorn}, which are essential for HRI, by asking the follower to fulfill the leader's requirements first and then searching for the leader to resume the person-following behavior. Some third-view and robot-egocentric visual examples are shown in Figure~\ref{fig:thirdViews} and Figure~\ref{fig:teaserImage}, respectively. 

As a result, we compiled a large-scale, long-term, multi-modal, and diverse egocentric dataset consisting of 35 sequences captured in different locations, with multiple recordings taken at various times (e.g., daytime, sunset, and night). Some multi-modal examples are shown in Figure~\ref{fig:datasetExamples}.

\subsection{Annotation and Labels} \label{sec:label}
Our dataset includes annotations for the target person in panoramic images with 2D bounding boxes. In the following, we will discuss the details of the annotation procedure, bounding box parameterization and additional metadata.

The annotation is performed by all involved students, who are specially instructed for the task. Considering the target's irregular motion and complex backgrounds, we annotate frames at 15 Hz for accuracy. We adopt an annotation standard similar to LaSOT~\citep{fan2021lasot} that if the target person appears in a frame, a labeler should ensure the bounding box is the tightest up-right one to fit any visible part of the target person; otherwise, the labeler gives an absent label, indicating the target person is under full occlusions or invisible. 

Since our goal is to annotate a target person, we leverage the capabilities of existing people detection and tracking models to alleviate the annotation workload. First, we apply all ``MOT + Target-ReID'' methods to a given sequence and visually inspect the resulting tracking videos to select the best tracking result as a prior annotation. This prior annotation includes all bounding boxes tracked by the MOT model, as well as those corresponding to the identified target, based on Target-ReID.

\begin{figure}[t]
        \centering
        \includegraphics[width=\linewidth]{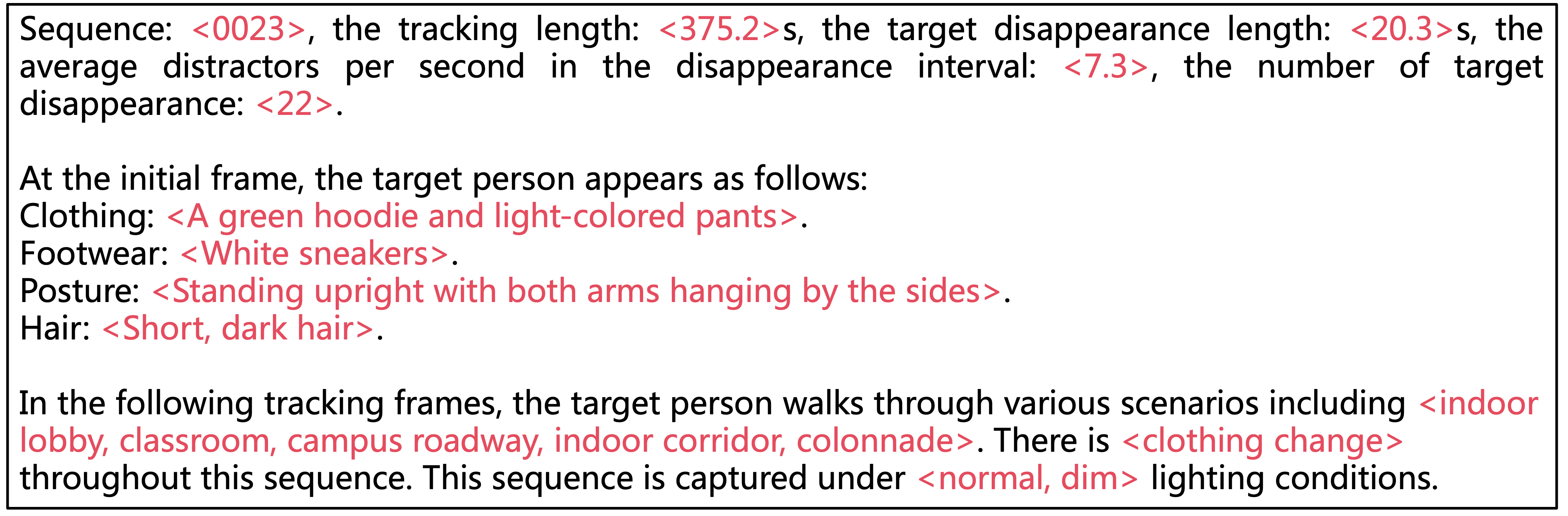}
\caption{\textbf{The description example.} The annotated description involves the basic attributes, the target person's appearance at the initial frame, the scenarios involved, the lighting conditions, and whether a clothing change occurs.}
\label{fig:descriptionDemo}
\end{figure}

\begin{figure}[t]
        \centering
        \includegraphics[width=\linewidth]{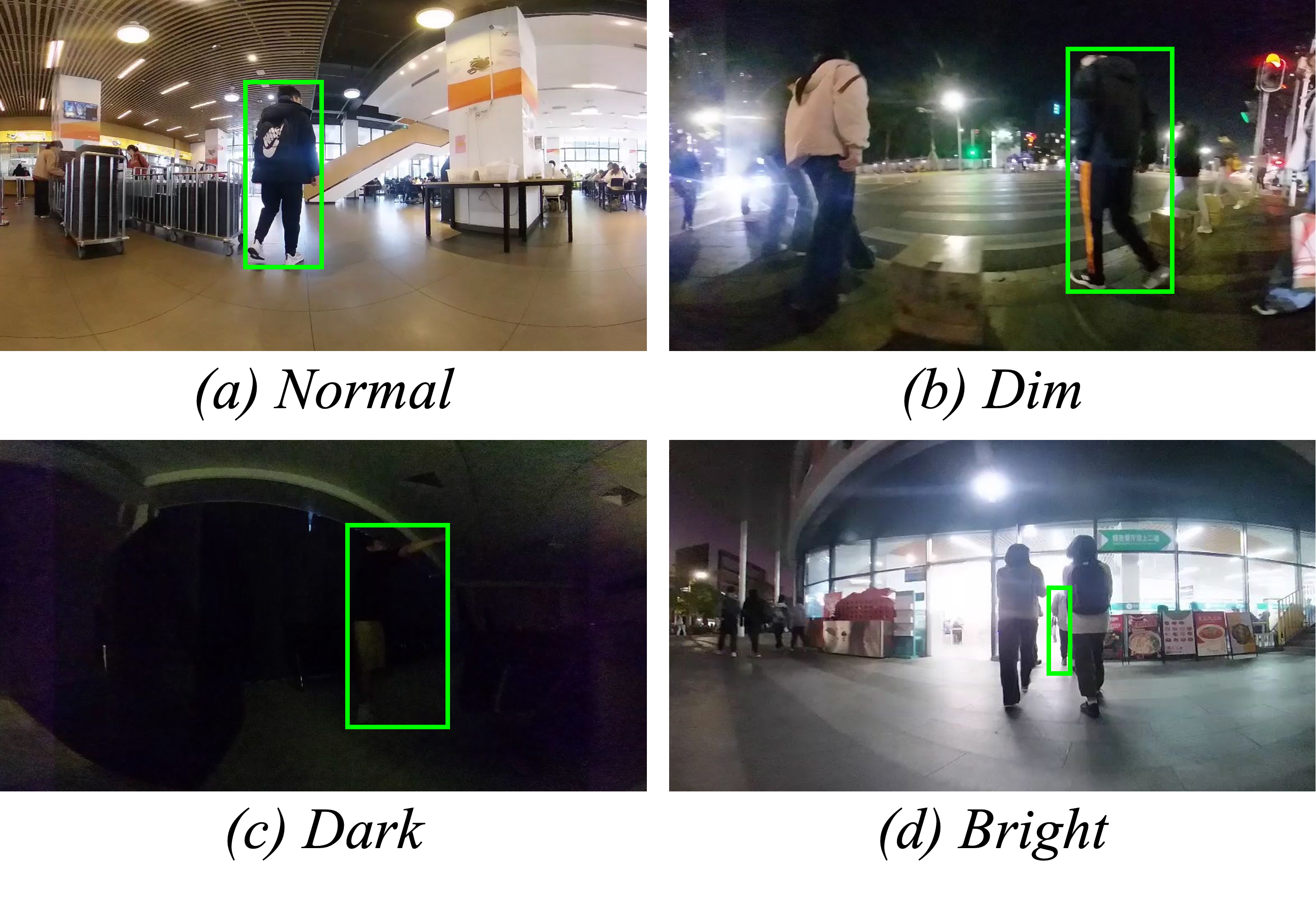}
\caption{\textbf{Examples of different lighting conditions.} (a) Normal: Well-lit environment with adequate visibility. (b) Dim: Low light conditions with reduced clarity. (c) Dark: Extremely low or no visible lighting. (d) Bright: Overexposed lighting, causing high brightness and potential loss of detail.}
\label{fig:lightCategory}
\end{figure}

Next, the annotator reviews each frame, refining any imperfectly estimated bounding boxes if the target is present. If the target does not exist visually, an ``absent'' label is applied. In cases where the target exists but is misclassified as absent by the Target-ReID method, we first identify the target by selecting its ID from the tracked bounding boxes. Then, we either fine-tune the estimated bounding box or manually draw a new one if no bounding box for the target is available. Finally, we upsample the annotations from 15 Hz to 30 Hz using linear interpolation. For each frame, we record the following metadata (most of them are followed by JRDB~\citep{martin2021jrdb}):
\begin{itemize}
    \item \texttt{is\_exist}: A binary indicator where $1$ signifies the target's presence and $0$ indicates the target is absent.
    \item \texttt{is\_behind\_glass}: An indicator that specifies whether the target is behind glass, where $-1$ denotes the absence of the target, $1$ indicates the target is behind the glass, and $0$ means the target is not behind it. This annotation is especially useful in indoor environments with glass walls or transparent obstacles, as depth sensors may incorrectly estimate the target's distance.
    \item \texttt{bbox}: A 2D bounding box is represented as $[u, v, w, h]$, where $(u, v)$ denotes the coordinates of the upper-left corner, and $(w, h)$ represents the width and height of the box. If the target is absent, the bounding box is denoted as $[0, 0, 0, 0]$.
    \item \texttt{interpolated}: A flag indicating whether the labels in this frame are directly derived from human annotations (at 15 Hz) or generated through interpolation of human-annotated frames (interpolated to 30 Hz).
\end{itemize}
Additionally, for each sequence, we provide textual descriptions detailing the basic attributes, the target person's appearance in the initial frame, the scenarios involved, the lighting conditions, and whether a clothing change occurs. An example of such a description is shown in Figure~\ref{fig:descriptionDemo}. The appearance description is initially generated by GPT-4o~\citep{openai2025gpt4o} and subsequently refined by a human annotator. Regarding the scenario, annotators review the video, subjectively select a frame that best represents the current unique scenario within the sequence, and submit it to GPT-4o with the prompt, ``Where is this environment? Use some words to represent it.'' The most appropriate word to describe the scenario is then chosen. The standards for categorizing lighting conditions are provided in Figure~\ref{fig:lightCategory}.

\begin{figure}[t]
        \centering
        \includegraphics[width=\linewidth]{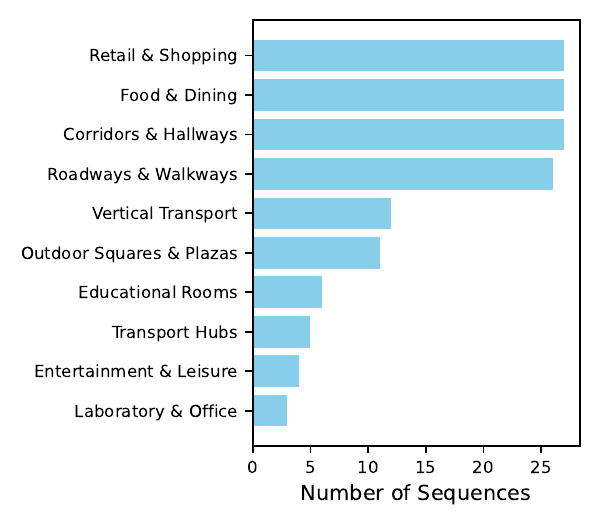}
\caption{\textbf{The distribution of scenario category counts.} Each sequence may include multiple scenarios that the target person walks through. This count reflects the number of sequences that involve each scenario category.}
\label{fig:categoryDistribution}
\end{figure}

\begin{figure}[t]
        \centering
        \includegraphics[width=\linewidth]{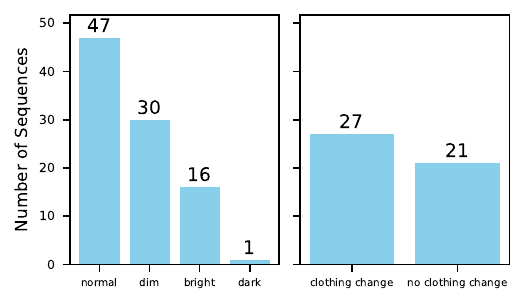}
\caption{\textbf{The distribution of lighting and clothing change conditions.} In the left subfigure, each sequence may involve multiple lighting conditions, with the count reflecting the number of sequences associated with each lighting condition. In the right subfigure, the clothing change condition is binary, where the count indicates the number of sequences in which a clothing change occurs or does not occur.}
\label{fig:lightAndClothes}
\end{figure}

\begin{figure}[t]
        \centering
        \includegraphics[width=\linewidth]{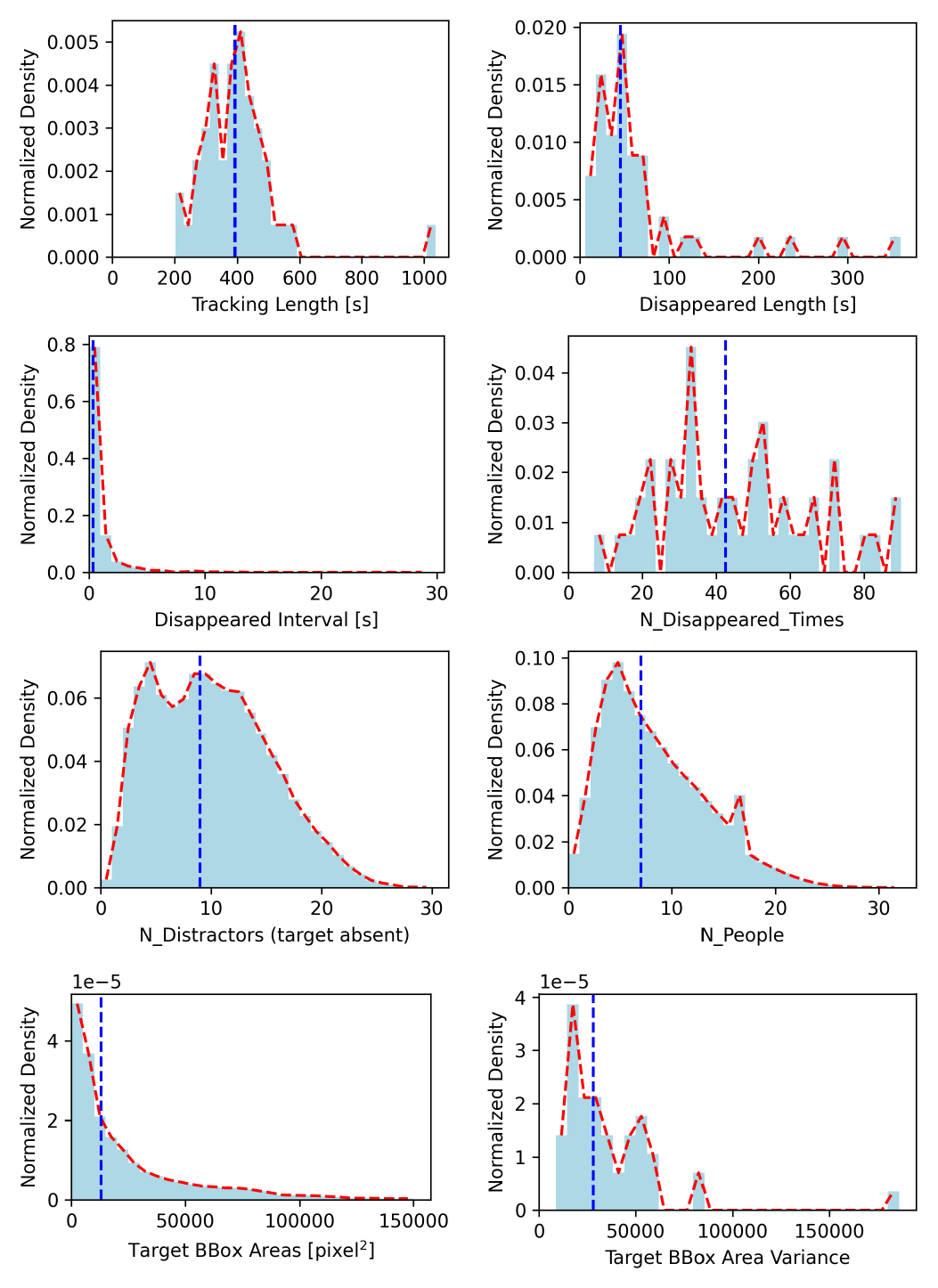}
\caption{\textbf{The histogram of dataset statistics.} In the ``disappeared intervals'' subfigure, durations over 30 seconds are excluded since they are considered part of the search procedure. The blue line represents the median.}
\label{fig:statistics}
\end{figure}

\begin{figure}[t]
        \centering
        \includegraphics[width=0.95\linewidth]{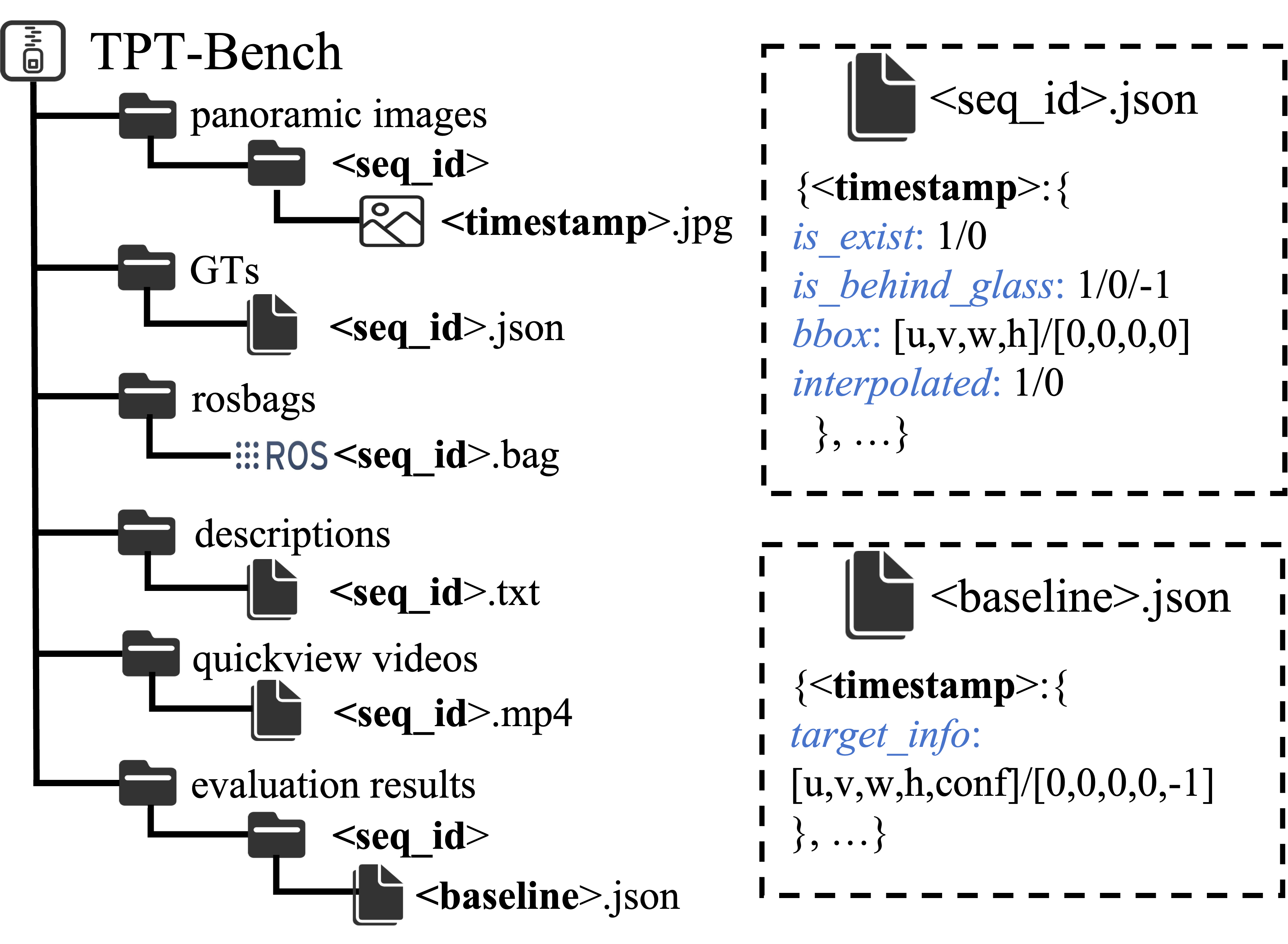}
\caption{\textbf{The dataset organization.}}
\label{fig:dataOrg}
\end{figure}

\subsection{Dataset Statistics} \label{sec:statistics}
In this benchmark, we assess the difficulties that long-term TPT algorithms encounter across diverse real-world situations. The overall distribution of scenario categories is visualised in Figure~\ref{fig:categoryDistribution}. These ten high-level categories are distilled from the full set of labelled scenarios listed in Table~\ref{tab:scenarios} (Appendix C). Most recordings take place in environments where people typically conduct everyday activities, such as \textit{retail \& shopping}, \textit{food \& dining}, \textit{corridors \& hallways}, and \textit{roadways \& walkways}. In addition, the sequences purposely span diverse lighting conditions and frequent clothing changes by the target person (Figure~\ref{fig:lightAndClothes}), both of which alter visual appearance and substantially increase the challenge for long-term TPT.

Table~\ref{tab:datasets} contrasts key tracking statistics of our dataset with those of other person-tracking benchmarks, with most metrics sourced from~\cite{lukezivc2020performance}. Figure \ref{fig:statistics} then visualises the distributions of sequence duration, disappearance duration and interval, total disappearance count, distractor count during absences, the number of surrounding people, target bounding-box area, and the intra-sequence variance of that area.

Compared with existing datasets, ours offers markedly longer videos, with a mean sequence duration of 397.2$\pm$123.9 seconds. These extended recordings capture numerous disappearances: the target is absent for 65.3 s on average per sequence, occasionally exceeding 200 s---conditions akin to challenging person-search tasks. Each sequence contains 45.4 disappearances on average (maximum 90), with most gaps lasting 0–6 s, indicating frequent occlusions followed by rapid re-emergence. While the target is missing, the sensor’s field of view still contains a mean of 9.6 distractors and sometimes more than 20, underscoring the difficulty of re-identification in crowded scenes. Finally, the median target bounding-box area is roughly 114.3 $\text{pixel}^2$, with a per-sequence variance of around 167.4 $\text{pixel}^2$, reflecting substantial scale changes due to unrestricted movement---another key obstacle for robust long-term tracking.

\subsection{Dataset Organization} \label{sec:dataOrg}
Figure~\ref{fig:dataOrg} illustrates the organisation of our dataset. All sensor streams are captured as ROS bag files, leveraging ROS’s mature support for multi-modal visualisation, debugging and precise time synchronisation. Within the \textit{panoramic images} folder, each sequence directory (\texttt{$<$seq\_id$>$}) saved this sequence's panoramic images, named as \texttt{$<$timestamp$>$.jpg}. Parallel folders hold the ground-truth annotations (\textit{GTs}), natural-language sequence descriptions (\textit{descriptions}), baseline tracker outputs (\textit{evaluation results}), and the raw ROS bag files themselves (\textit{rosbags}). For rapid inspection, a low-resolution, cropped and annotated video of every sequence is also provided in \textit{quickview videos}.

 
\begin{figure}[t]
        \centering
        \includegraphics[width=\linewidth]{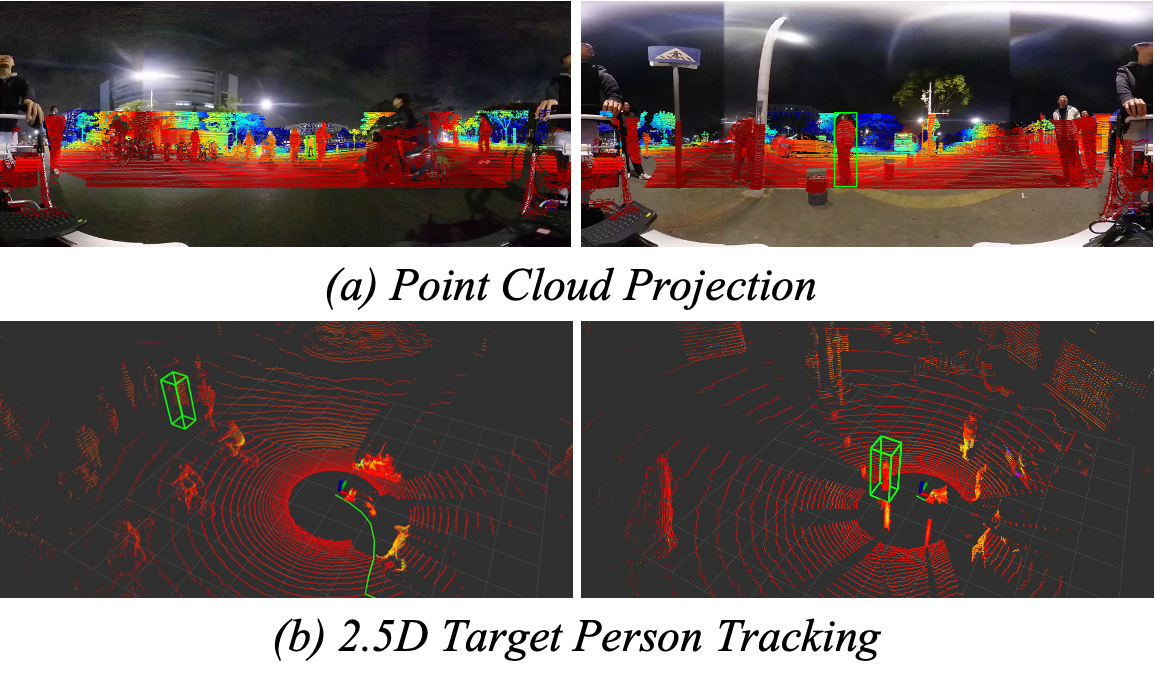}
\caption{\textbf{Development tool usages.} (a) The projected point cloud overlaid onto panoramic images using our custom development tools, with point colors encoding relative distances. This process involves calibration, annotation loaders, and point cloud processing. (b) The 2.5D ground-plane tracking of the tracked target person. This is computed by 1) using bounding box annotations to extract the target’s median depth from the projected point cloud, 2) estimating the target’s position from the panoramic projection, and 3) applying a Kalman Filter to smooth the resulting trajectory.}
\label{fig:develTools}
\end{figure}

\subsection{Development Tools} \label{sec:develTools}
We have released a comprehensive suite of tools that enable users to evaluate their TPT algorithms using our dataset. In addition, we provide detailed instructions for running SOTA methods on the dataset, along with the evaluated results for all baseline methods. To further facilitate the utilization of multi-modal data, we offer calibration parameters and a set of scripts based on \cite{wei2024fusionportablev2}, which include functions for loading calibration parameters, synchronizing and storing ROS messages, extracting and visualizing odometry paths, and projecting point clouds onto pinhole cameras. We have also extended several functionalities, such as panoramic camera projection and unprojection, panoramic image distortion, and target person tracking in 2.5D space. Examples of point cloud projection on panoramic images and target person tracking are shown in Figures~\ref{fig:develTools} (a) and~\ref{fig:develTools} (b), respectively. The current tracking accuracy is limited by partial occlusions, which cause depth estimation biases; ongoing improvements will further enhance the capabilities of this toolkit.

\subsection{Ethical Considerations}
This dataset was collected by continuously recording a consenting individual moving through public spaces such as cafeterias and shopping malls. Explicit informed consent was obtained from the primary subject before data collection.
We acknowledge that other individuals may have been incidentally captured in the footage while appearing in the background in these public areas. Given that the data was collected in publicly accessible spaces where individuals generally do not have a reasonable expectation of privacy, and no personally identifiable information (PII) of bystanders was intentionally recorded or used, we believe that the collection adheres to ethical guidelines. This study received ethical approval from the Southern University of Science and Technology IRB (approval \#2025JSJ128) on April 01, 2025.

\section{Experiment} \label{sec:experiment}
In this section, we first present the evaluation protocol and metrics from the perspectives of video object tracking and robotics applications (Sec.~\ref{sec:metric}). Next, we introduce baselines that are capable of addressing TPT problems (Sec.~\ref{sec:baseline}). We then analyze and discuss the experimental results on the TPT-Bench (Sec.~\ref{sec:sot} and Sec.~\ref{sec:expTargetReID}). Finally, we evaluate trackers on different datasets to demonstrate the TPT-Bench is more challenging with the objective of long-term tracking. (Sec.~\ref{sec:expDifferentDataset}).

\subsection{Evaluation Protocol and Metrics} \label{sec:metric}
This dataset can be divided into training and testing subsets; however, in this study, we use all 48 sequences to assess tracking performance, providing a large-scale evaluation of the trackers. Using the initial template, we evaluate tracking performance over sequential frames. Since we target the evaluation of long-term tracking ability, we adopt widely-used metrics in long-term object tracking~\citep{lukezivc2020performance}: \textit{tracking precision} (\textit{TP}), \textit{tracking recall} (\textit{TR}) and \textit{F-score} (\textit{F}). 

For each frame at timestep $t$, we have $G_t$ as the ground-truth target bounding box, $A_t(\tau_{\theta})$ as the estimated bounding box by the tracker, $\theta_t$ the prediction confidence and $\tau_{\theta}$ a classification threshold. The ground truth is empty if the target person disappears, i.e., $G_t = \emptyset$. Similarly, the estimation is empty, i.e., $A_t(\tau_{\theta})=\emptyset$ if the prediction confidence is below a classification threshold, i.e., $\theta_t < \tau_{\theta}$. The agreement between the ground truth and the prediction is quantified using the intersection-over-union (IoU), denoted as $\Omega(A_t(\tau_{\theta}), G_t)$. By definition, this IoU is zero if either $A_t(\tau_{\theta})$ or $G_t$ is empty. A true positive is then determined by whether the overlap $\Omega(A_t(\tau_{\theta}), G_t)$ exceeds a predefined IoU threshold $\tau_{\Omega}$. Given the two thresholds $(\tau_{\theta}, \tau_{\Omega})$, \textit{TP} and \textit{TR} are defined as:
\begin{equation}
\begin{split}
    &{\textit{TP}}(\tau_{\theta}, \tau_{\Omega}) = |\{t: \Omega(A_t(\tau_{\theta}), G_t) \geq \tau_{\Omega} \cap A_t(\tau_{\theta}) \neq \emptyset\}|/N_p \\
    &{\textit{TR}}(\tau_{\theta}, \tau_{\Omega}) = |\{t: \Omega(A_t(\tau_{\theta}), G_t) \geq \tau_{\Omega} \cap G_t \neq \emptyset\}|/N_g
\end{split}
\end{equation}
where $|\cdot|$ is the cardinality, $N_g$ is the number of frames with $G_t \neq \emptyset$, and $N_p$ is the number of frames with target prediction, i.e., $A_t(\tau_{\theta}) \neq \emptyset$. To quantify the accuracy of target absence prediction and target ReID capabilities, \cite{lukezivc2020performance} re-formulates the above equations by integrating over $\tau_{\Omega}$, resulting in ${\textit{TP}}(\tau_{\theta})$ and ${\textit{TR}}(\tau_{\theta})$. Then \textit{F} is defined as:
\begin{equation}
    \textit{F}(\tau_{\theta}) = 2{\textit{TP}}(\tau_{\theta}){\textit{TR}}(\tau_{\theta})/({\textit{TP}}(\tau_{\theta})+{\textit{TR}}(\tau_{\theta}))
\end{equation}
On this basis, the primary \textit{F-score} is defined as the highest F-score at an optimal threshold.

Additionally, we include max recall at 100\% precision (\textit{MR}), a crucial metric~\citep{yu2024gv} for the robotics community often overlooked in existing object tracking benchmarks. This metric evaluates the tracking performance with zero false positives, as false positives are particularly detrimental for robotics systems, where even a false positive could mistakenly lead the robot to follow the wrong person. From the detection literature, the definition of a true positive is tricky since it depends on the IoU threshold $\tau_{\Omega}$. To solve this, for example, \cite{vcehovin2016visual} averages performance over various thresholds. Similarly, we average \textit{MR} over various IoU thresholds. As consequence, average max recall at 100\% precision (\textit{AMR}) is defined as: 
\begin{equation}
    \textit{AMR} = \frac{\sum_{\tau_{\Omega}}^{\{\tau_{\Omega}\}} \max \{\textit{TR}(\tau_{\theta},\tau_{\Omega}) | \textit{TP}(\tau_{\theta},\tau_{\Omega})=1, \tau_{\theta}\ \in [0,1] \}}{|\{\tau_{\Omega}\}|}
\end{equation}

Moreover, we adopt the short-term tracking metric~\citep{vcehovin2016visual}, average overlap (\textit{AO}) for evaluating the tracking accuracy over all frames that the target person is visible, i.e., \textit{TR} at $\tau_{\theta}=0$.

\subsection{Baselines} \label{sec:baseline}
We present a systematic evaluation of long-term tracking capabilities in existing TPT methods using our proposed dataset. Given the diversity of TPT approaches, we strategically select exemplary high-performing representatives from three fundamental paradigms: \textbf{SOT}, \textbf{MOT + Target-ReID}, and \textbf{Detection + ReID}.

For SOT, we build upon the baseline comparisons in \cite{tang2024egotracks} and include a range of trackers to comprehensively evaluate long-term tracking performance. Our selection encompasses short-term trackers with online target representation, such as DiMP~\citep{dimp}, SiameseRPN++~\citep{li2019siamrpn++}, and TAMOs~\citep{tamo}. Additionally, we evaluate state-of-the-art short-term transformer-based trackers, including ToMP (ResNet101)~\citep{tomp}, STARK~\citep{stark}, and MixFormer variants (ConvMAE~\citep{mixformer2022cvpr} and CVT~\citep{mixformer2023tpami}). For specialized long-term tracking, we incorporate LTMU~\citep{ltmu}, which integrates a global tracker (GlobalTrack~\citep{huang2020globaltrack}) with a local tracker (DiMP), and Siam-RCNN~\citep{siamrcnn}, which leverages dynamic programming to model historical target paths. We also evaluate KeepTrack~\citep{keeptrack}, a distractor-aware tracker designed for crowded scenarios, to assess its suitability for TPT tasks.

For ``MOT + Target-ReID,'' we adopt ByteTrack~\citep{zhang2022bytetrack} combined with three contemporary Target-ReID methods: CARPE-ID~\citep{rollo2024icra}, RPF-ReID~\citep{ye2024person}, and RPF-ReID with online continual learning (OCL)~\citep{ye2024person}. CARPE-ID employs an adaptive target feature updating strategy based on exponential moving averages, while RPF-ReID trains an online ridge-regression classifier using target and non-target features. RPF-ReID with OCL further enhances performance by fine-tuning the feature extractor with online-identified samples (including both target and non-target images), thereby adapting to long-term appearance changes. To investigate the impact of person-ReID capability on TPT performance, we utilize two feature extractors: ResNet18 (R18) pre-trained on MOT20~\citep{dendorfer2020mot20} and KPR~\citep{kpr}, a transformer-based partial-ReID method pre-trained on the OCC-Duke dataset~\citep{miao2019pose}. The latter is particularly well-suited for occlusion-rich, robot-egocentric scenarios. Notably, because KPR’s training process relies on human joint annotations—which are challenging to obtain during OCL—we do not apply OCL to this feature extractor.

Finally, we implement a ``Detection + ReID'' baseline as a conventional pipeline. We use a fixed-size queue to store target features and identify the target person by selecting the detected candidate (by YOLOX~\citep{ge2021yolox}) with the highest feature similarity. This approach mirrors standard practices in ReID-driven MOT systems and serves as a foundational comparison point for more advanced methods. As with the above baselines, we adopted two feature extractors (R18 and KPR) for evaluation. Method flows of the above baselines are provided in the supplementary materials.

\begin{table}[t]
    \centering
    \caption{\textbf{Tracker performance comparison on the TPT-Bench.} Off-the-shelf, all trackers perform poorly, demonstrating the new challenges of TPT-Bench. Higher performance from tracking by the type of ``ByteTrack~\citep{zhang2022bytetrack}+Target-ReID.'' The best result is indicated in bold, while the second-best result is underscored.}
    \scalebox{0.9}{
    \begin{tabular}{lcccc}
        \toprule
        \bf Method &\bf AO &\bf F-Score &\bf AMR &\bf FPS \\
        \midrule
        \midrule
        \makecell[l]{DiMP\\\citep{dimp}} &11.40 &10.58 &0.00 &53.9 \\
        \makecell[l]{MixFormer-CVT\\\citep{mixformer2022cvpr}} &12.73 &13.52 &0.37 &35.9 \\
        \makecell[l]{SiameseRPN++\\\citep{li2019siamrpn++}} &11.52 &14.46 &1.68 &55.4 \\
        \makecell[l]{MixFormer-ConvMAE\\\citep{mixformer2023tpami}} &20.04 &19.49 &0.12 &18.9 \\
        \makecell[l]{TAMOs\\\citep{tamo}} &21.75 &22.55 &0.02 &12.9 \\
        \makecell[l]{ToMP\\\citep{tomp}} &29.64 &30.31 &0.00 &39.1 \\
        \makecell[l]{LTMU\\\citep{ltmu}} &37.11 &37.04 &2.29 &13.1 \\
        \makecell[l]{KeepTrack\\\citep{keeptrack}} &34.67 &37.15 &3.59 &28.6 \\
        \makecell[l]{STARK\\\citep{stark}} &38.84 &39.21 &0.89 &36.4 \\
        \makecell[l]{Siam-RCNN\\\citep{siamrcnn}} &41.07 &40.05 &7.71 &3.7 \\
        \midrule
        \makecell[l]{Detection w/ R18\\\citep{dendorfer2020mot20}} &38.00 &35.78 &3.31 &20.7 \\
        \makecell[l]{Detection w/ KPR\\\citep{kpr}} &56.97 &54.56 &6.41 &10.2 \\
        \midrule
        \multicolumn{5}{l}{\textbf{ByteTrack~\citep{zhang2022bytetrack}\ +\ Target-ReID}} \\
        CARPE-ID w/ R18 &21.46 &20.72 &0.62 &21.2 \\
        \makecell[l]{CARPE-ID w/ KPR\\\citep{rollo2024icra}} &56.41 &55.28 &9.84 &12.0 \\
        
        RPF-ReID w/ R18 &48.38 &46.87 &3.89 &20.3 \\
        \makecell[l]{RPF-ReID w/ KPR\\\citep{ye2024person}} &\bf 67.44 &\bf 66.06 &\bf 16.26 &11.5 \\
        \makecell[l]{RPF-ReID+OCL w/ R18\\\citep{ye2024person}} &\underline{60.25} &\underline{58.58} &\underline{11.33} &18.5 \\
        
        \bottomrule
    \end{tabular}}
    \label{tab:results}
\end{table}

\begin{table}[t]
    \centering
    \caption{\textbf{Tracker performance on different datasets.} Compared to existing benchmarks, the trackers on the TPT-Bench exhibit significantly lower performance in terms of F-Score and AMR, highlighting more challenging long-term tracking scenarios.}
    \label{tab:resultsOnDifferentDatasets}
    \scalebox{0.65}{
    \begin{tabular}{lcccccc}
        \toprule
        \multirow{2}{*}{\bf Method} & \multicolumn{2}{c}{\bf \makecell[c]{Chen's dataset\\\citep{chen2017crv}}} & \multicolumn{2}{c}{\bf \makecell[c]{LaSOT\\\citep{fan2021lasot}}} & \multicolumn{2}{c}{\bf \makecell[c]{TPT-Bench\\}} \\
            & \textit{F-Score} & \textit{AMR} & \textit{F-Score} & \textit{AMR} & \textit{F-Score} & \textit{AMR} \\
        \midrule
        \midrule
        \makecell[l]{STARK\\\citep{stark}} & \underline{97.8} & 54.0 & \bf 74.3 & \underline{32.0} & 39.2 & 0.9 \\
        \makecell[l]{Siam-RCNN\\\citep{siamrcnn}} & 94.3 & \bf 65.9 & 69.0 & 30.2 & 40.1 & 7.7 \\
        \makecell[l]{Detection w/ KPR\\\citep{kpr}} & 94.4 & 37.4 & 58.5 & 22.9 & 54.6 & 6.4 \\
        \makecell[l]{RPF-ReID w/ KPR\\\citep{ye2024person}} & \bf 99.1 & \underline{55.5} & \underline{70.3} & \bf 37.1 & \bf 66.1 & \bf 16.3 \\
        \makecell[l]{RPF-ReID+OCL w/ R18\\\citep{ye2024person}} & 94.4 & 47.9 & 65.6 & 25.8 & \underline{58.6} & \underline{11.3} \\
        \bottomrule
    \end{tabular}}
\end{table}

\subsection{Performance of SOT trackers} \label{sec:sot}
Baseline results on the TPT-Bench are presented in Table~\ref{tab:results}, and several key observations are worth highlighting. First, long-term trackers demonstrate superior performance, with high F-Scores for LTMU (37.04), KeepTrack (37.15), and Siam-RCNN (40.05). In contrast, most short-term trackers struggle on the TPT-Bench, achieving lower F-Scores, such as DiMP (10.58), MixFormer-ConvMAE (19.49), TAMOs (22.55), and ToMP (30.31). This result is expected as long-term trackers are designed with a focus on re-detection. An exception among short-term trackers is STARK, which achieves the highest F-Score at 39.21, owing to its second training stage that teaches the model to classify the presence of the object.

Second, while STARK achieves a high F-Score, it reports a relatively low AMR of 0.89. In contrast, long-term trackers exhibit higher AMRs, such as LTMU (2.29), KeepTrack (3.59), and Siam-RCNN (7.71). This suggests that a well-designed re-detection mechanism is crucial for the identification and tracking of the target person---an essential capability in robotics, where incorrect person identification can lead to undesired behavior.

Finally, we observe that SOT methods generally underperform on the TPT-Bench when compared to the top-performing methods from other types that leverage prior knowledge about humans. For instance, while Siam-RCNN achieves a higher AMR than ``Detection w/ KPR'' (+1.30), its F-Score is lower by -14.51. Additionally, its F-Score and AMR are both lower than ``CARPE-ID w/ KPR'' by -15.23 and -2.13, ``RPF-ReID w/ KPR'' by -26.01 and -8.55, and ``RPF-ReID+OCL w/ R18'' by -18.53 and -3.62. This is because SOT methods typically focus on general object tracking and do not consider the use of human-related prior knowledge. Other types of methods incorporate prior knowledge about humans, such as people detection (spatial cues) and person ReID features (appearance cues), which can be more beneficial for TPT.

\subsection{Performance of Target-ReID Methods} \label{sec:expTargetReID}
Target-ReID refers to the methodology used to construct the appearance model of a target person. These appearance models usually leverage pre-trained ReID features. From Table~\ref{tab:results}, we can observe that each method equipped with KPR performs better than that with ResNet18. For example, ``Detection w/ KPR'' achieves 54.56 F-Score and 6.41 AMR, larger than ``Detection w/ R18'' by +18.78 F-Score and +3.10 AMR. This is a straightforward verification of the positive impact of ReID ability on TPT since the target person is directly identified based on a feature similarity matching. Furthermore, other methods also benefit from stronger ReID features, e.g., ``RPF-ReID w/ KPR'' achieves the best result with 66.06 F-Score and 16.26 AMR, larger than ``RPF-ReID w/ R18'' by +19.19 F-Score and +12.37 AMR.

Leveraging the same feature extractor, RPF-ReID outperforms CARPE-ID. For instance, when equipped with KPR, it surpasses CARPE-ID by +10.78 F-Score and +6.42 AMR; with R18, it exceeds CARPE-ID by +26.15 F-Score and +3.27 AMR. This improvement can be attributed to Ridge Regression, which optimizes the classification boundary by leveraging both target and non-target ReID features collected online. In contrast, CARPE-ID merely aggregates target ReID features in a decaying manner within the original ReID feature space, failing to enhance discrimination further.

Besides learning a classifier online, ``RPF-ReID+OCL w/ R18'' also fine-tunes the feature extractor using one of the replay-based OCL techniques, Reservoir~\citep{ER}. This method consolidates the memory by randomly deleting samples based on a probabilistic distribution related to observation times. It then replays training samples from the memory to fine-tune the feature extractor. As a result, it achieves 58.58 F-Score and 11.33 AMR, surpassing ``RPF-ReID w/ R18'' by +11.71 F-Score and +7.44 AMR.

\subsection{Performance on Different Datasets} \label{sec:expDifferentDataset}
To highlight the more challenging long-term tracking scenarios in TPT-Bench, we further evaluate several high-performing trackers on other datasets that also involve tracking a person over an extended period (LaSOT~\citep{fan2021lasot} and Chen's dataset~\citep{chen2017crv}). The results are shown in Table~\ref{tab:resultsOnDifferentDatasets}. It is evident that TPT-Bench poses greater long-term tracking challenges compared to the other two datasets, as the best performance achieved by a tracker is only 66.1 F-Score and 16.3 AMR. These values are significantly lower than those in LaSOT, with a drop of -8.2 F-Score and -20.8 AMR, and in Chen's dataset, with a drop of -33.0 F-Score and -49.6 AMR.

Furthermore, we observe that the best baseline evaluated in TPT-Bench, namely ``RPF-ReID w/ KPR,'' performs well in the other two person-tracking datasets as well. According to the F-Score metric, it achieves an impressive 99.1 F-Score in Chen's dataset and a second-best 70.3 F-Score in LaSOT. This is only -4.0 F-Score lower than the best-performing tracker, STARK~\citep{stark}, which is pre-trained on LaSOT. This demonstrates that leveraging prior knowledge of human appearance can significantly enhance the generalizability of TPT in diverse tracking scenarios.

\begin{figure*}[t]
        \centering
        \includegraphics[width=0.7\linewidth, trim=10 15 10 10, clip]{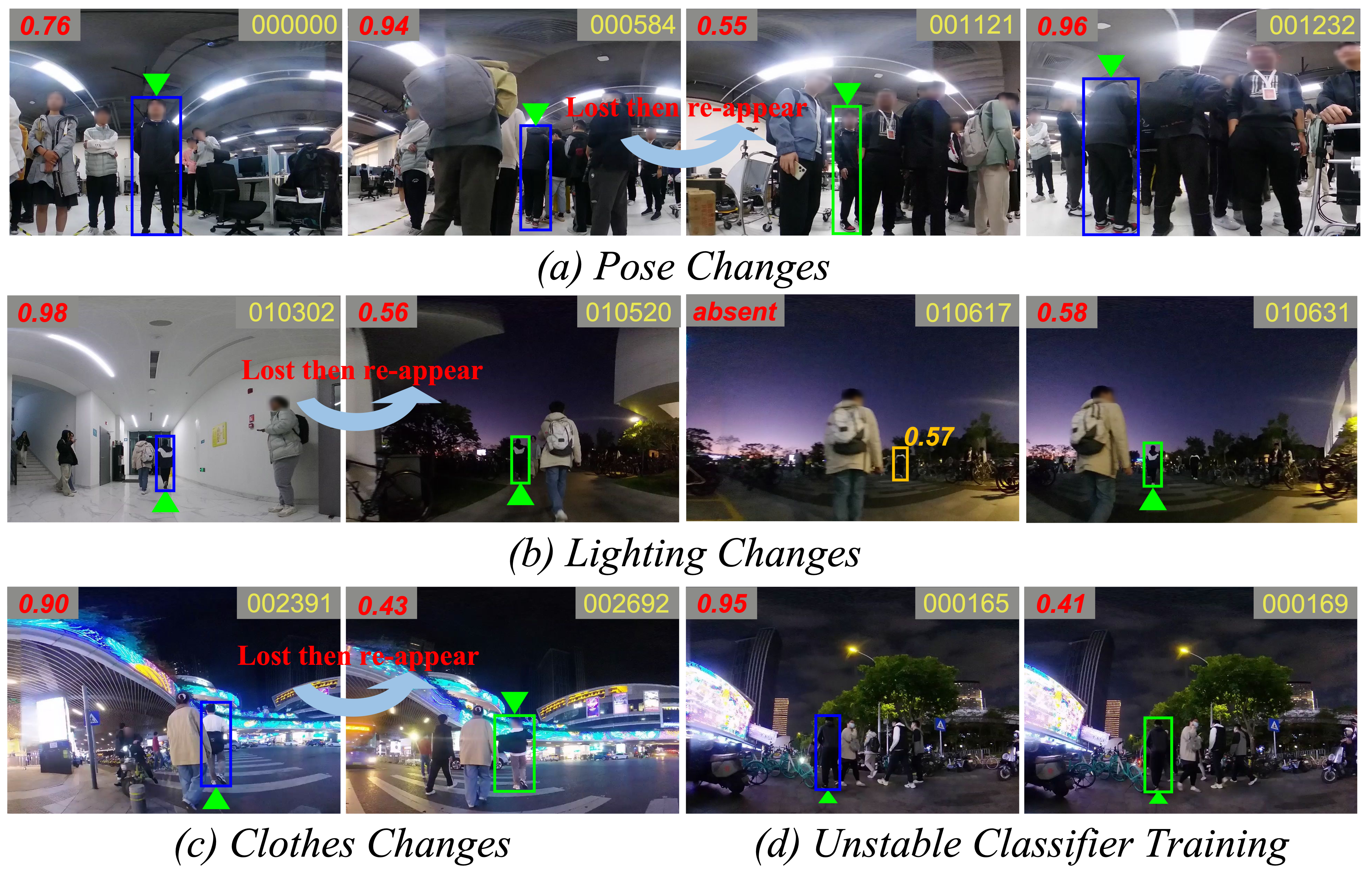}
\caption{\textbf{Failure cases.}  The upper-left number shows the estimated confidence of the target and the upper-right number shows the frame number. The green triangle and box represent the ground truth, while the blue box shows the estimation. For the third subfigure of (b), the target is absent; we mark the yellow box, representing the highest confidence estimation in this frame. The images are cropped to focus on the target person for clarity.}
\label{fig:failureCases}
\end{figure*}

\section{Discussion} \label{sec:discussion}
\subsection{Potential Improvements of TPT Algorithms}
The above experiments (Sec.~\ref{sec:experiment}) suggest ``RPF-ReID w/ KPR'' to be the most competitive method. We focus on this method for additional analysis, suggesting improvements to long-term egocentric TPT. From experiments, we identify three key takeaways for the potential improvement:

\subsubsection{\textbf{Designing better TPT-driven memory consolidation:}} A failure case shown in Figure~\ref{fig:failureCases} (a) illustrates a scenario where the method fails to re-identify the target from the front-view (third subfigure) until it reappears from the back-view (fourth subfigure). This failure under pose changes may stem from the classifier being trained primarily on abundant back-view observations of the target before it is lost (second subfigure). However, the front-view observation is present in earlier frames (first subfigure). A similar case is shown in Figure~\ref{fig:failureCases} (c), where the classifier fails to re-identify the target after slight clothing changes, even though the reappeared appearance had been observed in previous frames. This highlights the need for improved memory consolidation methods that capture diverse historical samples to better support target re-identification under appearance variation.

\subsubsection{\textbf{Improving feature extractor:}} Although there are no changes in pose or clothing in Figure~\ref{fig:failureCases} (b), lighting variations cause a significant drop in the estimated confidence when the target reappears, indicating insufficient lighting robustness in the pre-trained ReID model. Additionally, in the third subfigure of Figure~\ref{fig:failureCases}, where the target is absent, a distractor obtains a high score of 0.57, comparable to that of the true target. To avoid false positives, the classifier must produce high-confidence estimates for the target to reliably distinguish it from distractors. These challenges can be attributed to the non-discriminative representation of the pre-trained ReID features. To improve target discrimination with high confidence, integrating feature uncertainty learning~\citep{yu2019robust} into our previously evaluated state-of-the-art partial ReID method, KPR~\citep{kpr}, could be an option to mitigate the impact of noisy training samples. Furthermore, our dataset offers a valuable resource for training more discriminative ReID features.

\subsubsection{\textbf{Stablizing classifier training:}} In Figure~\ref{fig:failureCases} (d), despite minimal visual changes, performance suddenly degrades, with estimated confidence dropping from 0.95 to 0.41. This illustrates the stability-plasticity dilemma~\citep{mai2022online}, where the classifier struggles to maintain stable representations while incorporating new samples. While many methods~\citep{mai2022online} address this challenge, they have been primarily evaluated on simplified datasets. Our dataset presents a realistic scenario, and adapting existing solutions to the TPT task remains an open question.

\subsection{Why Pushing A Cart for Data Collection}
Over the last decade, several datasets have been proposed to enhance robots' ability to navigate in human environments~\citep{martin2021jrdb,karnan2022socially,nguyen2023toward}. In addition to JRDB~\citep{martin2021jrdb} that highlights perception challenges, some datasets~\citep{karnan2022socially,nguyen2023toward} aim to teach robots to navigate socially in human-populated environments. For example, the SCAND dataset~\citep{karnan2022socially} includes sequential robot sensor data collected through teleoperated demonstrations to learn socially aware and independent navigation behaviors in human-populated settings. While teleoperated demonstrations can convey human intentions, they suffer from delays in navigation decision-making and are not convenient for non-robotics researchers. This is particularly critical in the person-following scenarios, where we need to respond quickly, considering both the target person and the surrounding environment.

In addition to teleoperated demonstrations, the MuSoHu dataset~\citep{nguyen2023toward} collects data using a sensor-equipped helmet worn by a naturally walking human. However, this method overlooks the inherent kinematic constraints of robots and the nuances of realistic robot-centric observation, which typically involve increased occlusion and obstruction scenarios encountered in real-world deployments. While the aforementioned data collection methodologies effectively capture egocentric observations and the corresponding human decision-making processes—thereby aiding the development of social navigation techniques—we propose an alternative approach. Specifically, our TPT dataset is collected by having a human push a sensor-equipped cart while following, accompanying, or occasionally searching for a leading person. This method not only reflects natural person-following behavior in human-populated environments but also considers the actual kinematic constraints of robotic platforms.

More recently, a study~\citep{wang2025trackvla} has shown the feasibility of training an end-to-end Vision-Language-Action (VLA) model for robot person following, utilizing collected pairs of image observations and local trajectories. While effective, the data in their study was collected in a simulation with manually rendered distractors walking randomly. Furthermore, their experiments only demonstrate the potential for human following in simple scenarios, without considering social rules, long-term person search, or target re-identification in crowded environments. We believe our dataset will not only serve as a valuable benchmark for robot person following at the perception level but will also advance research in imitation learning, contributing to the development of more robust VLA models.

\subsection{Limitations and Future Work}
Building a task-driven robotics dataset that spans diverse, human-populated scenarios is inherently labor-intensive. To capture the challenges of long-term TPT, we intentionally oversampled a large pool of sequences and then curated the most varied subset for annotation. While this approach yields a rich benchmark, the dataset still has notable limitations. Here, we detail these constraints and outline potential future work, aiming to guide and inspire subsequent research in the field.

\subsubsection{\textbf{Multi-Modal Annotation and Tracking:}}
Although our release provides rich multi-modal sensor streams and a suite of development tools, it currently lacks annotations for certain modalities---most notably LiDAR point clouds. Such a modality is particularly valuable for long-term TPT, especially in environments with severe lighting changes. A promising direction for future work is to exploit the existing 2D bounding box annotations, together with calibrated cross-sensor extrinsics, to facilitate the semi-automatic generation of accurate 3D bounding boxes. As discussed in Sec.~\ref{sec:develTools}, partial occlusions pose a significant challenge to precise depth estimation of the target. One potential solution is to incorporate vision foundation models, such as SAM2~\citep{ravi2024sam}, to obtain more precise target masks, thereby improving depth estimation. By combining these refined depth estimates with the motion constraints of the target person, it is possible to optimize the sequence of 3D bounding box positions. Human annotators can then verify and refine these results to produce high-quality annotations. These additional labels would enable the construction of a dedicated multi-modal TPT benchmark and foster further research in this direction.

\subsubsection{\textbf{Social-Aware Data Collection and Annotation:}}
In this dataset, demonstrators are instructed to walk freely in human-populated environments, highlighting the challenges of long-term tracking. The follower's behavior inherently involves implicit social interactions, such as maintaining appropriate distance or exhibiting politeness toward nearby pedestrians. However, the dataset lacks other common forms of implicit social behaviors, such as comfortable approach and engagement~\citep{repiso2018robot}, or group following~\citep{repiso2020people}. An important direction for future work is to collect socially-aware demonstrations encompassing these interactions, and to annotate them with corresponding language instructions.

Moreover, the dataset currently does not include explicit interactions—namely, direct communication between the target person and the follower. In many real-world scenarios, users may issue commands to the robot to adjust its position or speed based on personal preferences. Common communication modalities include hand gestures~\citep{rollo2023followme} and vocal commands~\citep{yoshimi2006development}. A promising extension would be to annotate the target person’s 2D skeletal joints and incorporate audio recordings via onboard microphones to capture spoken instructions.

\subsubsection{\textbf{Dataset Diversity and Scale:}}
Developing reliable and socially aware human-following robots is gaining increasing importance in human-centric environments. To facilitate this, comprehensive datasets or benchmarks are essential for training and evaluating human-following algorithms. In principle, the diversity of a benchmark determines the upper bound of algorithmic performance, as demonstrated by the recent advances in foundation models for computer vision and natural language processing~\citep{ravi2024sam,depth_anything_v1,liu2024deepseek}. However, collecting large-scale, diverse, real-world datasets remains a significant challenge. TPT-Bench addresses a key gap in existing benchmarks by providing long-term, egocentric tracking data captured in crowded, real-world environments. These settings feature frequent occlusions, dynamic obstructions, and variations in lighting and scale. 

Nonetheless, we acknowledge certain limitations in the current version of the dataset in terms of its scale and diversity: (1) \textbf{Scenarios:} The dataset lacks coverage of critical environments such as factories and hospitals, where appearance-similar uniforms pose challenges for target re-identification. Moreover, these environments are often rich in context, requiring human-following robots to exhibit contextually appropriate behavior~\citep{francis2025principles,honig2018toward}. (2) \textbf{Complex Pose Variations:} The dataset predominantly features walking sequences and does not include complex human motions such as dancing, fitness exercises, parkour, bicycling, or running. These motions are particularly challenging for continuous tracking yet represent common use cases for human-following robots. (3) \textbf{Scale:} The dataset is relatively small, with a limited number of sequences. A larger dataset would enable the establishment of train/test splits and foster the development of learning-based methods tailored specifically to the TPT task. TPT-Bench represents a foundational effort toward enabling socially aware human following. As the research community expands, we envision the creation of more diverse, large-scale datasets that will drive further advancements in this domain.

\section{Conclusion}
In this paper, we introduce TPT-Bench, the first large-scale, multi-modal, robot-egocentric dataset for target person tracking in diverse, crowded and unstructured environments. In addition to challenges in general tracking benchmarks, e.g.,  partial occlusion, large distortion, changes of clothing, scale and lighting, and motion blur, the TPT-Bench highlights long-term tracking challenges, including frequent target disappearance and the need for target re-identification among numerous pedestrians.

Using this dataset, we conduct comprehensive experiments to assess the performance of state-of-the-art trackers, revealing that they struggle with low F-scores and a maximum recall of 100\% precision. Among the tested trackers, those integrating MOT and target-ReID outperform traditional SOT trackers, likely due to the incorporation of human-related prior knowledge, such as person detection and ReID features. By releasing this dataset, we believe it can encourage further research and applications in target-person-tracking, benefiting communities of computer vision, robotics and HRI.


\bibliographystyle{SageH}
\bibliography{ref}
\section*{Appendix A: Method flows of evaluated TPT algorithms}
In this paper, we evaluate three types of trackers that are capable of performing TPT, including \textbf{SOT}, \textbf{MOT+Target-ReID}, and \textbf{Detection+ReID}. Given only the initial location of the target person's bounding box, SOT trackers~\citep{dimp, li2019siamrpn++, tamo, tomp, stark, mixformer2022cvpr, mixformer2023tpami, ltmu, siamrcnn, keeptrack} infer estimated bounding box and its confidence of sequent frames with deep neural networks. Due to the large number of SOT methods evaluated in this paper, providing pseudocode for each approach is infeasible within the scope of this work. The pipeline of ``Detection+ReID'' is shown in Algorithm~\ref{alg:detectionReID}, where a queue is employed to store high-confidence target features and identify the target person by selecting the detected candidate (by YOLOX~\citep{ge2021yolox}) with the highest feature similarity. 

For the type of ``MOT+Target-ReID,'' we evaluate different target-ReID methods based on the MOT results from ByteTrack~\citep{zhang2022bytetrack}. In this paper, we evaluate ``CAPRE-ID~\citep{rollo2024icra},'' ``RPF-ReID~\citep{ye2024person},'' and ``RPF-ReID+OCL~\citep{ye2024person}.'' The pipeline of ``CAPRE-ID'' is illustrated in Algorithm~\ref{alg:carpeid}. In this approach, the target feature is treated as a distribution, and a damped exponential moving average method is employed to update this distribution. This technique helps update the target appearance model gradually, preventing overfitting to newly observed, similar target features.

\begin{algorithm}[t]
    \DontPrintSemicolon
    \SetNoFillComment
    \footnotesize
    \KwIn{A video sequence \texttt{V}, Initial target bounding box $B_{\text{init}}$; object detector \texttt{Det}; feature extractor \texttt{Ext}, target feature queue \texttt{Queue}, insertion confidence threshold $\tau$}
    \KwOut{Target bounding box $B$ and the confidence $c$}

    \For{$frame$ $I_k$ in \texttt{V}}{
        \uIf {$k==0$} {
            $feat=\texttt{Ext}(I_k,B_{\text{init}})$ \\
            \texttt{Queue} $\leftarrow feat$ \\
            \textbf{Return} $\{B_{\text{init}},1.0\}$
        }
        \uElse{
            $\mathcal{D}_k=$\texttt{Det}($f_k$) \\
            $\mathcal{T}_k \leftarrow \emptyset$ \\
            \For{$d$ in $\mathcal{D}_k$}{
                $feat=\texttt{Ext}(I_k,d)$ \\
                $sim = \frac{\sum_{i=1}^N feat \cdot \texttt{Queue}[i]}{N}$ \\
                $\mathcal{T}_k \leftarrow \mathcal{T}_k \cup \{d,feat,sim\}$ \\
            }
            $\{d_{\text{max}},feat_{\text{max}},sim_{\text{max}}\} \leftarrow \max(\mathcal{T}_k[sim])$ \\
            \uIf {$sim_{\text{max}}>\tau$} {
                \texttt{Queue} $\leftarrow feat_{\text{max}}$ \\
            }
            
        }
        \textbf{Return} $\{d_{\text{max}},sim_{\text{max}}\}$
    }
\caption{Detection+ReID}
\label{alg:detectionReID}
\end{algorithm}

\begin{algorithm}[t]
    \DontPrintSemicolon
    \SetNoFillComment
    \footnotesize
    \KwIn{A video sequence \texttt{V}, Initial target bounding box $B_{\text{init}}$; person tracker \texttt{Track}; feature extractor \texttt{Ext}, target feature $feat_{\text{tar}}$, damped version of the exponential moving average \texttt{DEMA}, refer to CARPE-ID~\citep{rollo2024icra}.}
    \KwOut{Target bounding box $B$ and the confidence $c$}
    \For{$frame$ $I_k$ in \texttt{V}}{
        $\{d,id\}_k = \texttt{Track}(I_k)$ \\
        $feats=\texttt{Ext}(I_k,\{d\}_k)$ \\

        \uIf {$k==0$} {
            $id_{\text{tar}}=\arg\max \texttt{IoU}(B_{\text{init}},\{d\}_k)$ \\
            $feat_{\text{tar}}=feats[id_{\text{tar}}]$ \\
            
            \textbf{Return} $\{B_{\text{init}},1.0\}$
        }
        \uElse{
            \uIf{$id_{\text{tar}} \in \{id\}_k$}{
                \textcolor{blue}{\# Updating target appearance model} \\
                $sim_{\text{max}} = \texttt{CalCosSim}(feat_{\text{tar}},feats[id_{\text{tar}}])$ \\
                $feat_{\text{tar}},\lambda_s=\texttt{DEMA}(feat_{\text{tar}},feats[id_{\text{tar}}])$ \\
                \textbf{Return} $\{\{d\}_k[id_{\text{tar}}], sim_{\text{max}}\}$ \\
                
            }
            \uElse{
                \textcolor{blue}{\# Target re-identification} \\
                $sims = \texttt{CalCosSim}(feats,feat_{\text{tar}})$ \\
                $id_{\text{max}} = \arg \max sims$ \\
                $sim_{\text{max}}=sims[id_{\text{max}}], d_{\text{max}}=\{d\}_k[id_{\text{max}}]$ \\
                \uIf {$sim_{\text{max}}>\lambda_s$}{
                    $id_{\text{tar}}=id_{\text{max}}$ \\
                }
                \textbf{Return} $\{d_{\text{max}},sim_{\text{max}}\}$
            }
        }
    }
\caption{CARPE-ID}
\label{alg:carpeid}
\end{algorithm}

RPF-ReID~\citep{ye2024person} treats the target appearance model as a classifier, training it with features collected online. Ridge regression is employed for classification. A simplified pipeline is presented in Algorithm~\ref{alg:rpfreid}. In addition to classifier training, they introduce the fine-tuning of the feature extractor using a replay buffer. This buffer stores online-collected image patches and corresponding labels (target and non-target). A key design element is the consolidation mechanism, driven by replay-based OCL techniques. This mechanism enhances the diversity of stored samples. In the evaluation, we employ the Reservoir~\citep{ER} method for consolidation. The resulting approach is called ``RPF-ReID+OCL,'' with a simplified diagram shown in Algorithm~\ref{alg:rpfreidocl}.

Note that when using part-based ReID features, such as KPR \citep{kpr}, which can infer part-based features and part visibility, we construct multiple part-based target appearance models for target confidence estimation. The final confidence is then obtained by averaging all individual estimations. All methods are evaluated on a PC with an Intel® Core™ i9-12900K CPU and NVIDIA GeForce RTX 3090.

\begin{table}[t]
    \centering
    \caption{\textbf{Performance comparison of different target-ReID methods on the TPT-Bench.} The best result is indicated in bold, while the second-best result is underscored.}
    \scalebox{0.75}{
    \begin{tabular}{lcccc}
        \toprule
        \bf Method &\bf AO &\bf F-Score &\bf AMR &\bf FPS \\
        \midrule
        \midrule
        \multicolumn{5}{l}{\textbf{ByteTrack~\citep{zhang2022bytetrack}\ +\ Target-ReID}} \\
        CARPE-ID w/ R18 &21.46 &20.72 &0.62 &21.2 \\
        \makecell[l]{CARPE-ID w/ KPR\\\citep{rollo2024icra}} &56.41 &55.28 &9.84 &12.0 \\
        RPF-ReID w/ R18 &48.38 &46.87 &3.89 &20.3 \\
        \makecell[l]{RPF-ReID w/ KPR\\\citep{ye2024person}} &\underline{67.44} &\underline{66.06} &\bf 16.26 &11.5 \\
        
        \makecell[l]{RPF-ReID+OCL w/ R18\\\citep{ye2024person}} &60.25 &58.58 &\underline{11.33} &18.5 \\
        \makecell[l]{RPF-ReID+OCL w/ Parts-ResNet18\\\citep{ye2024person}} &\bf 68.11 &\bf 68.54 &8.15 &7.09 \\
        \bottomrule
    \end{tabular}}
    \label{tab:partOCL}
\end{table}

\begin{table}[t]
\centering
\caption{\textbf{Detailed annotated scenarios, merged categories.}}
\scalebox{0.76}{
\begin{tabular}{ll}
\toprule
\textbf{Category} & \textbf{Scenarios} \\
    \midrule
    \midrule
    \textbf{\makecell[l]{Retail \\\& Shopping}} & \makecell[l]{bakery, book store, bookstore, clothing store, \\ distribution center, gift store, retail store \\ mobile phone store, supermarket, shopping mall, \\ shopping mall walkway, pickup point, photo booth} \\
    \midrule
    \textbf{\makecell[l]{Food \\\& Dining}} & \makecell[l]{student canteen, canteen, food court, \\ coffee bar, street food market, restaurant, \\ fast-food restaurant, Chinese restaurant, \\ McDonald’s, KFC, tea shop} \\
    \midrule
    \textbf{\makecell[l]{Corridors \\\& Hallways}} & \makecell[l]{research institute hallway, school hallway, \\ campus hallway, indoor corridor, indoor lobby, \\ connected corridor, classroom corridor, \\ campus corridor, corridor, hallway, \\ breezeway, colonnade, lobby} \\
    \midrule
    \textbf{\makecell[l]{Roadways \\\& Walkways}} & \makecell[l]{campus roadway, campus, covered walkway, \\ footbridge, sidewalk, pedestrian walkway, \\ pedestrian crosswalk, urban walkway, \\ street crossing, crosswalk, zebra crosswalk} \\
    \midrule
    \textbf{\makecell[l]{Vertical Transport}} & \makecell[l]{elevator, escalator} \\
    \midrule
    \textbf{\makecell[l]{Outdoor Squares \\\& Plazas}} & \makecell[l]{square, outdoor plaza, campus courtyard} \\
    \midrule
    \textbf{Educational Rooms} & \makecell[l]{classroom, lecture hall} \\
    \midrule
    \textbf{Transport Hubs} & \makecell[l]{bus stop, subway station, metro} \\
    \midrule
    \textbf{\makecell[l]{Entertainment \\\& Leisure}} & \makecell[l]{game center, game store, pool hall, \\ amusement center} \\
    \midrule
    \textbf{\makecell[l]{Laboratory \\\& Office}} & \makecell[l]{robotics laboratory, meeting room, office} \\
    \bottomrule
\end{tabular}}
\label{tab:scenarios}
\end{table}

\section*{Appendix B: RPF-ReID+OCL with human joints}
According to \cite{ye2024person}, an enhanced approach involves leveraging human joint information to explicitly represent partially visible human bodies. Using estimated visibilities, they transform local features from the feature extractor (R18) into features associated with specific body parts. Additionally, part-based classifiers are developed for target confidence estimation. The results, referred to as "RPF-ReID+OCL w/ parts-R18," are presented in Table~\ref{tab:partOCL}, alongside results from other target-ReID methods for comparison.

We observe a significant improvement in F-Score by +7.86, although the AMR decreases by -3.18 compared to "RPF-ReID+OCL w/ R18." Furthermore, the FPS drops to 7.1 due to additional computation of human joint estimation and the fine-tuning of the part-based feature extractor.

\begin{algorithm}[t]
    \DontPrintSemicolon
    \SetNoFillComment
    \footnotesize
    \KwIn{A video sequence \texttt{V}, Initial target bounding box $B_{\text{init}}$; person tracker \texttt{Track}; feature extractor \texttt{Ext}, target classifier \texttt{Cls}, feature queue $\mathbb{S}$, ID switch threshold $\tau_{\text{sw}}$, ReID threshold $\tau_\text{re}$.}
    \KwOut{Target bounding box $B$ and the confidence $c$.}
    \For{$frame$ $I_k$ in \texttt{V}}{
        $\{d,id\}_k = \texttt{Track}(I_k)$ \\
        $feats=\texttt{Ext}(I_k,\{d\}_k)$ \\

        \uIf {$k==0$} {
            $id_{\text{tar}}=\arg\max \texttt{IoU}(B_{\text{init}},\{d\}_k)$ \\
        }
        \uIf{$id_{\text{tar}} \in \{id\}_k$}{
            \textcolor{blue}{\# Updating target appearance model} \\
            $\mathbb{S} \leftarrow \{feats[id_{\text{tar}}], 1\}$ \\
            \For{$i$ in $\{id\}_k\setminus{id_{\text{tar}}}$}{
                $\mathbb{S} \leftarrow \{feats[i], 0\}$ \\
            }
            Sample $m$ from $\mathbb{S}$, train \texttt{Cls} with $m$ \\
            $sim_{\text{max}} = \texttt{Cls}(feats[id_{\text{tar}}]), d_{\text{max}}=\{d\}_k[id_{\text{tar}}]$ \\
            \uIf{$sim_{\text{max}}<\tau_{\text{sw}}$}{
                $id_{\text{tar}}=-1$ \\
            }
            \textbf{Return} $\{d_{\text{max}}, sim_{\text{max}}\}$ \\
        }
        \uElse{
            \textcolor{blue}{\# Target re-identification} \\
            $sims = \texttt{Cls}(feats)$ \\
            $id_{\text{max}} = \arg \max sims$ \\
            $sim_{\text{max}}=sims[id_{\text{max}}], d_{\text{max}}=\{d\}_k[id_{\text{max}}]$ \\
            \uIf {$sim_{\text{max}}>\tau_\text{re}$}{
                $id_{\text{tar}}=id_{\text{max}}$ \\
            }
            \textbf{Return} $\{d_{\text{max}},sim_{\text{max}}\}$
        }
    }
\caption{RPF-ReID}
\label{alg:rpfreid}
\end{algorithm}

\begin{algorithm}[t]
    \DontPrintSemicolon
    \SetNoFillComment
    \footnotesize
    \KwIn{A video sequence \texttt{V}, Initial target bounding box $B_{\text{init}}$; person tracker \texttt{Track}; feature extractor \texttt{Ext}, target classifier \texttt{Cls}, feature queue $\mathbb{S}$, replay buffer $\mathbb{L}$, ID switch threshold $\tau_{\text{sw}}$, ReID threshold $\tau_\text{re}$.}
    \KwOut{Target bounding box $B$ and the confidence $c$.}
    \For{$frame$ $I_k$ in \texttt{V}}{
        $\{d,id\}_k = \texttt{Track}(I_k)$ \\
        $feats=\texttt{Ext}(I_k,\{d\}_k)$ \\
        $patches=\texttt{crop}(I_k,\{d\}_k)$ \\
        \uIf {$k==0$} {
            $id_{\text{tar}}=\arg\max \texttt{IoU}(B_{\text{init}},\{d\}_k)$ \\
        }
        \uIf{$id_{\text{tar}} \in \{id\}_k$}{
            \textcolor{blue}{\# Updating target appearance model} \\
            $\mathbb{S} \leftarrow \{feats[id_{\text{tar}}], 1\}$, $\mathbb{L} \leftarrow \{patches[id_{\text{tar}}], 1\}$ \\
            \For{$i$ in $\{id\}_k\setminus{id_{\text{tar}}}$}{
                $\mathbb{S} \leftarrow \{feats[i], 0\}$, $\mathbb{L} \leftarrow \{patches[i], 0\}$ \\
            }
            Sample $m$ from $\mathbb{S}$, train \texttt{Cls} with $m$ \\
            Consolidate $\mathbb{L}$ with replay-based OCL \\
            Sample $l$ from $\mathbb{L}$, train \texttt{Ext} with $l$ \\
            $sim_{\text{max}} = \texttt{Cls}(feats[id_{\text{tar}}]), d_{\text{max}}=\{d\}_k[id_{\text{tar}}]$ \\
            \uIf{$sim_{\text{max}}<\tau_{\text{sw}}$}{
                $id_{\text{tar}}=-1$ \\
            }
            \textbf{Return} $\{d_{\text{max}}, sim_{\text{max}}\}$ \\
            }
        \uElse{
            \textcolor{blue}{\# Target re-identification} \\
            $sims = \texttt{Cls}(feats)$ \\
            $id_{\text{max}} = \arg \max sims$ \\
            $sim_{\text{max}}=sims[id_{\text{max}}], d_{\text{max}}=\{d\}_k[id_{\text{max}}]$ \\
            \uIf {$sim_{\text{max}}>\tau_{\text{re}}$}{
                $id_{\text{tar}}=id_{\text{max}}$ \\
            }
            \textbf{Return} $\{d_{\text{max}},sim_{\text{max}}\}$
        }
    }
\caption{RPF-ReID+OCL}
\label{alg:rpfreidocl}
\end{algorithm}

\section*{Appendix C: Annotated scenarios}
Ten high-level categories are distilled from the full set of labelled scenarios listed, which are shown in Table~\ref{tab:scenarios}.

\end{document}